%% file: naic.tex
\documentclass[10pt,journal,compsoc]{IEEEtran}

\ifCLASSOPTIONcompsoc
\usepackage[nocompress]{cite}
\else
\usepackage{cite}
\fi

\ifCLASSINFOpdf
\else
\fi

\usepackage{amsmath}
\usepackage{amsthm}

\usepackage{booktabs}

\usepackage{graphicx}

\usepackage{multirow}
\usepackage{makecell}
\usepackage{amssymb}
\usepackage{xspace}
\makeatletter
\DeclareRobustCommand\onedot{\futurelet\@let@token\@onedot}
\def\@onedot{\ifx\@let@token.\else.\null\fi\xspace}
\def\eg{\emph{e.g}\onedot} 
\def\ie{\emph{i.e}\onedot} 
 
\def\etc{\emph{etc}\onedot}

\makeatother

\usepackage{xcolor}

\makeatletter
\def\@IEEEsectpunct{.\ \,}
\def\paragraph{\@startsection{paragraph}{4}{\z@}{1.5ex plus 1.5ex minus 0.5ex}%
	{0ex}{\normalfont\normalsize\sffamily\bfseries}}
\def\italicparagraph{\@startsection{paragraph}{4}{\z@}{1.5ex plus 1.5ex minus 0.5ex}%
	{0ex}{\normalfont\normalsize\textit}}
\makeatother

\definecolor{shadecolor}{RGB}{150,150,150}

\usepackage{xcolor, soul}
\sethlcolor{gray}

\usepackage{boondox-calo}
\usepackage{subcaption}

\usepackage{balance}

\begin{document}
	\title{Fast Sequence Generation \\with Multi-Agent Reinforcement Learning}
	
	\author{Longteng Guo,~
		Jing Liu,~\IEEEmembership{Member,~IEEE,}
		Xinxin Zhu,~ %
		and Hanqing Lu,~\IEEEmembership{Senior~Member,~IEEE}
		\IEEEcompsocitemizethanks{\IEEEcompsocthanksitem Longteng Guo, Jing Liu, Xinxin Zhu and Hanqing Lu are with the National Laboratory of
			Pattern Recognition, Institute of Automation, Chinese Academy of Sciences,
			Beijing 100190, and also with the School of Artificial Intelligence, University 
			of Chinese Academy of Sciences, Beijing 100049, China. E-mail: \{longteng.guo, jliu, xinxin.zhu, luhq\}@nlpr.ia.ac.cn. 
			(Corresponding author: Jing Liu)
			\IEEEcompsocthanksitem This work has been submitted to the IEEE for possible publication. Copyright may be transferred without notice, after which this version may no longer be accessible.
		
		}%

	}

	\markboth{Journal of \LaTeX\ Class Files,~Vol.~14, No.~8, August~2015}%
	{Shell \MakeLowercase{\textit{et al.}}: Bare Demo of IEEEtran.cls for Computer Society Journals}

	\IEEEtitleabstractindextext{%
		\begin{abstract}
			Autoregressive sequence Generation models have achieved state-of-the-art performance in areas like machine translation and image captioning. These models are autoregressive in that they generate each word by conditioning on previously generated words, which leads to heavy latency during inference. Recently, non-autoregressive decoding has been proposed in machine translation to speed up the inference time by generating all words in parallel. Typically, these models use the word-level cross-entropy loss to optimize each word independently. However, such a learning process fails to consider the sentence-level consistency, thus resulting in inferior generation quality of these non-autoregressive models. In this paper, we propose a simple and efficient model for Non-Autoregressive sequence Generation (NAG) with a novel training paradigm: Counterfactuals-critical Multi-Agent Learning (CMAL). CMAL formulates NAG as a multi-agent reinforcement learning system where element positions in the target sequence are viewed as agents that learn to cooperatively maximize a sentence-level reward. On MSCOCO image captioning benchmark, our NAG method achieves a performance comparable to state-of-the-art autoregressive models, while brings 13.9$\times$ decoding speedup. On WMT14 EN-DE machine translation dataset, our method outperforms cross-entropy trained baseline by $6.0$ BLEU points while achieves the greatest decoding speedup of 17.46$\times$. 
		\end{abstract}
		
		}

	\maketitle

	\IEEEdisplaynontitleabstractindextext

	\IEEEpeerreviewmaketitle

	\IEEEraisesectionheading{\section{Introduction}\label{sec:introduction}}

	\IEEEPARstart{S}equence generation tasks aim at generating a target sequence conditioning on a source input. 
	Standard sequence generation tasks include neural machine translation \cite{wu2016google,vaswani2017attention}, image captioning \cite{vinyals2017show}, 
	and automatic speech recognition \cite{2007Automatic} \etc, where the targets are often sequences of words/tokens and the source inputs can be sentences, images, and speeches \etc. 
	Recent sequence generation models typically follow the encoder-decoder paradigm where an encoder encodes the input into vectorial representations,  
	and a sequence decoder, \eg recurrent neural networks (RNNs) or Transformer \cite{vaswani2017attention}, generates a sentence given the outputs of the encoder. 
	Most of these models use \textit{autoregressive} decoders that require sequential execution: %
	they generate each word conditioned on the sequence of previously generated words. 
	However, this process is not parallelizable and thus results in high inference latency, 
	which is sometimes unaffordable for real-time industrial applications. 
	
\begin{figure}[!t] 
		\centering
		\includegraphics[width=3.3in]{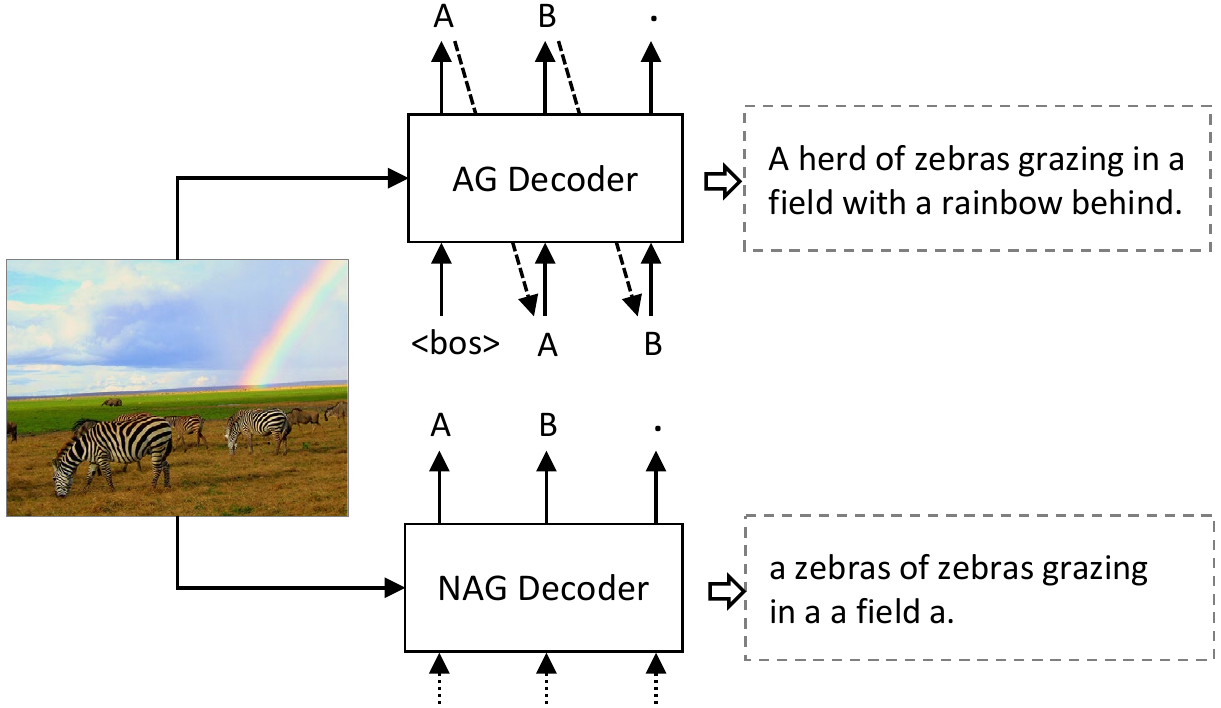}
		\caption{
			Given a source input, \eg an image, Autoregressive sequence Generation (AG) model generates a target sentence word by word, while 
			Non-Autoregressive sequence Generation (NAG) model outputs all words in parallel. 
		}
		\label{fig:first}
	\end{figure}

	Recently, {\em non-autoregressive} decoding was proposed in neural machine translation \cite{gu2017non} to significantly improve
	the inference speed by predicting all target words in parallel.  
	A non-autoregressive model takes basically the same structure as the autoregressive Transformer network \cite{vaswani2017attention}.  
	But instead of conditioning the decoder on the previously generated words as in autoregressive models, 
	they generate all words independently, as illustrated in Figure~\ref{fig:first}. 
	Such models are typically optimized with the cross-entropy (XE) losses of individual words. 
	
	However, existing non-autoregressive models still have a large gap in generation quality compared to their autoregressive counterparts, mainly due to their severe decoding inconsistency problem. 
	For example, in Figure~\ref{fig:first}, the caption generated by the non-autoregressive model has repeated words and incomplete content. 
	A major reason for such performance degradation is 
	that the word-level XE based training objective cannot guarantee sentence-level consistency. 
	That is, the XE loss encourages the model to generate the golden word in each position 
	but does not consider the global consistency of the whole sentence. 
	Also, though the models are trained with XE loss, however, at test time, they are typically evaluated using non-differentiable test metrics such as 
	CIDEr \cite{Vedantam2015CIDEr} or BLEU \cite{Papineni2002BLEU}. This inconsistency in training time and test time objectives is undesirable and can cause suboptimal performance. 
	
	To simultaneously reduce the inference time and improve the decoding consistency of sequence generation models,  
	in this paper, we propose a Non-Autoregressive sequence Generation (NAG) method with a novel training paradigm: 
	Counterfactuals-critical Multi-Agent Learning (CMAL). 
	We formulate NAG as a multi-agent reinforcement learning system, 
	where element positions in the target sequence are viewed as \textit{agents} and they act cooperatively to maximize a team reward about the quality of the whole sentence. 
	CMAL optimizes the model with policy gradient algorithm and introduces a \textit{counterfactual baseline}
	to address the multi-agent credit assignment \cite{chang2004all} problem. 
	Also, we further boost the generation quality by augmenting the training data with massive unlabeled inputs. 
	In the following, We introduce each component.  
	
	Specifically, our NAG model's architecture is based on Transformer \cite{vaswani2017attention} with the minimum necessary modifications. 
	We consider NAG decoder as a cooperative multi-agent reinforcement learning (MARL) \cite{bucsoniu2010multi} system. 
	In this system, each agent observes the \textit{environment} (the encoded visual context), 
	and communicates with other agents through the self-attention layers in Transformer. 
	After several rounds of environment observation and agent communication, 
	the agents reach an agreement about content planning of the target sentence and separately take \textit{actions} to predict the words in their corresponding positions, which forms a joint action. 
	The agents then receive a common sentence-level team \textit{reward} that is, for instance, the BLEU \cite{Papineni2002BLEU} score of the generated sentence.  
	The goal of the agents is to maximize the team reward and policy gradient is used to update their parameters. 
	This training paradigm has two benefits. 
	First, the non-differentiable test metrics, \eg BLEU \cite{Papineni2002BLEU} or CIDEr \cite{Vedantam2015CIDEr}, could be directly optimized. 
	Second, by optimizing the agents towards a common sentence-level objective, the decoding consistency can be substantially improved.

	A crucial challenge in the above MARL training paradigm is multi-agent credit assignment \cite{chang2004all}: 
	the shared team reward makes it difficult for each agent to deduce its own contribution to the team's success. 
	This could impede multi-agent learning and lead to decoding inconsistency. 
	To address this challenge, we compute an agent-specific advantage function that compares the team reward 
	for the joint action against an agent-wise \textit{counterfactual baseline} \cite{foerster2018counterfactual,chen2019counterfactual}. 
	The individual counterfactual baseline of an agent is the expected reward when marginalizing out a single agent's action, while keeping the other agents' actions fixed. 
	As a result, only actions from an agent that outperform the counterfactual baseline are given positive weight, 
	and inferior actions are suppressed.  
	To more precise credit assignment among agents, we further introduce a compositional counterfactual baseline as a supplement to the individual counterfactual baseline. 
	In the compositional counterfactual baseline, we treat two neighbouring agents as a \textit{compositional agent} to take into account the joint influence of their actions on the team reward. 
	The counterfactual baseline fully exploits the distinctive features of the multi-agent NAG system, \ie extremely short episode and large action space, and thereby is simple and easy to implement.  
	
	To further boost generation quality, we propose to augment the training data with massive unlabeled inputs. 
	These unlabeled data could be easily acquired without costly human annotations. 
	Specifically, we use an autoregressive teacher model to produce target sequences for these unlabeled inputs, 
	which are then used as pseudo paired data for training the non-autoregressive models.

	Compared to previous non-autoregressive models that often rely on specially designed submodules \cite{lee2018deterministic,pmlr-v80-kaiser18a}, 
	our NAG method adds almost no extra components to the Transformer architecture. 
	Owing to this structure simplicity, our model can maximize the decoding speedup. 
	We evaluate the proposed non-autoregressive sequence generation method on two distinctive sequence generation tasks, \ie image captioning and neural machine translation. 
	On MSCOCO image captioning benchmark, our method brings $13.9\times$ decoding speedup relative to the autoregressive counterpart, 
	while achieving comparable performance to state-of-the-art autoregressive models. 
	On the more challenging WMT14 EN-DE machine translation dataset, our method significantly outperforms the cross-entropy trained baseline by $6.0$ BLEU points while achieves a greatest decoding speedup of $17.46\times$.

	To summarize, the main contributions of this paper are four-fold: 
	
	\begin{itemize}
		\item  We propose a non-autoregressive sequence generation method with a novel training paradigm: 
		counterfactuals-critical multi-Agent learning. 
		To the best of our knowledge, we are the first to formulate non-autoregressive sequence generation 
		as a cooperative multi-agent learning problem. 
		\item We design individual and compositional counterfactual baselines to disentangle the contribution of each agent from the team reward. 
		\item We propose to utilize massive unlabeled data to boost the performance of non-autoregressive models. 
		\item Extensive experiments on image captioning and machine translation tasks show that 
		our method significantly outperforms previous cross-entropy trained non-autoregressive models while brings the greatest decoding speedup. 
	\end{itemize}

	\section{Related Work}
	\subsection{Autoregressive Sequence Generation}
	Sequence generation, \eg neural machine translation \cite{ranzato2015sequence,wu2016google} and image captioning \cite{xu2015show}, in deep learning has largely focused on autoregressive modeling. 
	These autoregressive models often follow the encoder-decoder framework with different choices of architectures such as RNN \cite{sutskever2014sequence,bahdanau2015neural}, convolutional neural networks (CNN) \cite{pmlr-v70-gehring17a} and full attention networks without recurrence and convolution (Transformer) \cite{vaswani2017attention}. 
	RNN-based models have a sequential architecture that prevents them from being parallelized during both training and test processes. 
	Although CNN and Transformer based models avoid recurrence at training time with highly parallelized architecture, the usage of autoregressive 
	decoding makes them still have to generate sequences token-by-token during inference. 
	Transformer has refreshed state-of-the-art performance on several sequence generation tasks,  including machine translation \cite{vaswani2017attention} and image captioning \cite{yu2019multimodal,guo2020normalized}.

	\subsection{Non-Autoregressive Sequence Generation}
	Non-autoregressive Sequence Generation (NAG) \cite{gu2017non} has recently been proposed to reduce the generation latency through parallelizing the decoding process. 
	A basic NAG model takes the same encoder-decoder architecture as Transformer.
	Non-autoregressive models often perform considerably worse than the autoregressive counterparts. 
	Some methods has been proposed to narrow the performance gap between autoregressive and non-autoregressive models, 
	including knowledge distillation \cite{gu2017non}, auxiliary regularization terms \cite{wang2019non}, 
	well-designed decoder inputs \cite{guo2019non}, 
	iterative refinement \cite{lee2018deterministic,gao2019masked} \etc. %
	Among them, \cite{gao2019masked} and \cite{fei2019fast} are the two published works on non-autoregressive image captioning. 
	However, these methods models typically use conventional word-level cross-entropy loss to optimize each word independently, which fails to consider the sentence-level consistency and results in poor generation quality. 
	Unlike these works, we propose to apply multi-agent reinforcement learning in non-autoregressive models to optimize a sentence-level objective.

	\subsection{Multi-Agent Reinforcement Learning (MARL)} 
	In a reinforcement learning system, agents take actions in an environment in order to maximize the cumulative reward.
	Some works on autoregressive sequence generation incorporated single-agent reinforcement learning to tackle the exposure bias problem \cite{ranzato2015sequence} and optimize non-differentiable score functions. 
	\cite{ranzato2015sequence} introduced REINFORCE \cite{Williams1992Simple} algorithm to sequence training with RNNs. 
	\cite{Rennie2016Self} lowers the variance of the policy gradient by introducing a baseline, which is the reward of the sequence generated by the inference algorithm. 
	\cite{bahdanau2016actor} propose actor-critic algorithms for sequence prediction. 
	However, applying reinforcement learning to autoregressive machine translation only reported marginal improvements \cite{wu2016google,wu-etal-2018-study,sun2020neural}. 
	Meanwhile, while previous works mainly focus on autoregressive models with single-agent reinforcement learning, 
	we study non-autoregressive models and are the first to formulate sequence generation models as a multi-agent reinforcement learning system. 
	
	Multi-Agent Reinforcement Learning (MARL) \cite{bucsoniu2010multi} considers a system of multiple agents that interact within a common environment. 
	It is often designed to deal with complex reinforcement learning problems that require decentralized policies, where each agent selects its own action. 
	Compared to well-studied MARL game tasks, our NAG model has a much larger action space   
	and extremely shorter episode.  
	Our counterfactual baseline gets intuition from \cite{foerster2018counterfactual}, which 
	requires training an additional critic network to estimate the Q value for each possible action. 
	Learning such a critic network increases the model complexity and is not practical due to the high-dimensional action space of NAG. 
	Instead, we turn to a simple yet powerful REINFORCE \cite{Williams1992Simple} algorithm in which the actual return is used to replace the Q function directly. 
	
	\subsection{Our Previous Work}
	Preliminary version of this work was published in \cite{ijcai2020-107}. 
	Compared with the previous version, this work shows two significant improvements.  
	First, we introduce \textit{compositional agents} to take into account the joint influence of the actions from two neighboring agents on the team reward. 
	With compositional agents, the compositional counterfactual baseline could be calculated as a supplement to the individual counterfactual baseline, 
	thereby promoting more precise credit assignment among agents.  
	Second, while in the previous version we only focus on image captioning task, 
	in this paper, we extend the model to more general sequence generation tasks and conduct extensive experiments on 
	classic sequence generation tasks including image captioning and machine translation. 
	Compared to image captioning task, machine translation task often has much longer target sentence length and is thus more challenging for non-autoregressive models. 
	We verify that our model, with only minor modifications, can perform quite well on non-autoregressive machine translation.  
	
	\section{Background}
	\subsection{The Transformer Model}
	
	Neural sequence generation models typically follow an encoder-decoder structure \cite{bahdanau2015neural, sutskever2014sequence, vinyals2017show}. 
	The encoder maps a source input to context representations. Given the context, the decoder then generates an output 
	sequence of tokens one token at a time. 
	The decoder is autoregressive \cite{graves2013generating}, consuming the generated partial sequence as input and output the next token.
	Transformer \cite{vaswani2017attention} is among the most successful neural sequence transduction models. 
	Both the encoder and decoder of Transformer are built with stacked multi-head attention and point-wise feed-forward layers. 
	
	\paragraph*{Multi-Head Attention} 
	At the core of Transformer is two types of multi-head attention, \ie 
	self-attention and inter-attention. 
	A general attention mechanism can be formulated as the weighted sum of the value vectors $V$ using the similarities between 
	query vectors $Q$ and key vectors $K$: 
	\begin{equation}
		\text{Attention}(Q, K, V)=\operatorname{softmax}\left(\frac{Q K^{T}}{\sqrt{d}}\right) \cdot V,
	\end{equation}
	where $d$ is the dimension of hidden representations. 
	For self-attention, $Q$, $K$ and $V$ are projected hidden representations of the preceding layer. 
	For inter-attention, $Q$ refers to projected hidden representations of the preceding layer, 
	whereas $K$ and $V$ are projected context vectors from the encoder outputs. 
	
	\paragraph*{Feed-Forward Network} 
	Position-wise feed-forward network is applied after each multi-head attentions. 
	It consists of a two-layer linear transformation with a ReLU activation in between:
	\begin{equation}
		\operatorname{FFN}(x)=\max \left(0, x W_{1}+b_{1}\right) W_{2}+b_{2}.
	\end{equation}
	
	Both the Transformer encoder and decoder compose of a stack of 6 identical blocks. 
	Each encoder block has two layers: a self-attention and a feed-forward network. 
	Each decoder block has three layers, including a self-attention, an inter-attention, and a feed-forward network. 
	Causal attention masking is applied in the self-attention of the decoder stack to prevent each position from attending to subsequent positions.

	\begin{figure*}[!t] 
		\centering
		\includegraphics[width=0.85\textwidth]{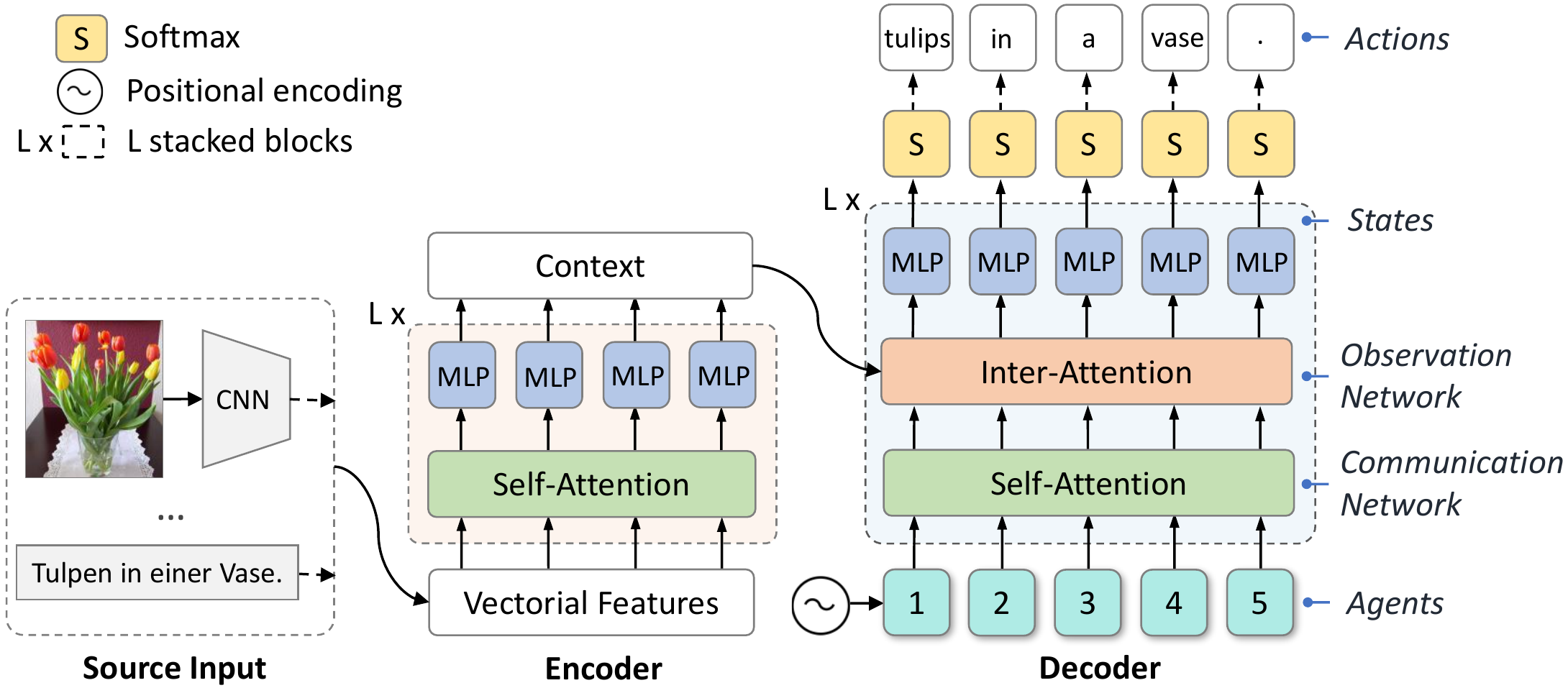}
		\caption{
			Illustration of our non-autoregressive sequence generation model. 
			It is based on Transformer and composes of an encoder and a decoder. 
			The encoder takes as input a source input, and the decoder generates a target sequence given the encoder outputs. 
			The source input could be, for example, the vectorial CNN features of an image in image captioning task or the token embeddings of a sentence in machine translation task. 
			On the rightmost, we cast the non-autoregressive decoder in the multi-agent reinforcement learning terminology, where  
			element positions in the target sequence are viewed as agents that learn to cooperatively maximize a sentence-level team reward. 
		}
		\label{fig:framework}
	\end{figure*}
	
	\subsection{Autoregressive Decoding}
	Given an input $x=(x_1, ..., x_n)$ and a target sentence $y=(y_1,..., y_T)$, Autoregressive sequence generation models 
	are based on a chain of conditional probabilities 
	with a left-to-right causal structure: 
	\begin{equation}
		p(y | I; \theta)=\prod_{i=1}^{T} p\left(y_{i} | y_{<i}, x; \theta \right),
	\end{equation}
	where $\theta$ is the model's parameters and $y_{<i}$ represents the words before the $i$-th word of target $y$. 
	The inference process is not parallelizable under such autoregressive factorization as 
	the sentence is generated word by word sequentially. 
	
	\subsection{Non-Autoregressive Decoding}
	Recently, non-autoregressive sequence models were proposed to alleviate the inference 
	latency by removing the sequential dependencies within the target sentence. 
	A NAG model generates all words independently:
	\begin{equation}
		p(y | I; \theta)=\prod_{i=1}^{T} p\left(y_{i} | x; \theta \right). 
		\label{eqn:na}
	\end{equation}
	During inference, all words could be parallelly decoded in one pass, 
	thus the inference speed could be significantly improved. 
	
	\subsection{Maximum Likelihood Training}
	Typically, a non-autoregressive sequence model straightforwardly adopts maximum likelihood training with a cross-entropy (XE) loss applied
	at each decoding position $i$ of the sentence: 
	\begin{equation}
		\mathcal{L}_{XE}(\theta)=-\sum_{i=1}^{T} \log \left(p\left(y_{i} | x; \theta \right)\right)
		\label{eqn:xe}
	\end{equation}
	However, such word-level XE loss only encourages the model to generate the golden word in each position, while the sentence-level consistency of the whole sentence is not modeled. 
	This problem becomes more serious in non-autoregressive decoding and causes non-autoregressive models to perform significantly poor than their autoregressive counterparts.

	\section{Approach} 
	In this section, we first present the architecture of our NAG model, 
	and then introduce our Counterfactuals-critical Multi-Agent Learning (CMAL) algorithm for model optimization. 
	Finally, we describe how we utilize unlabeled data to boost generation quality. 
	
	\subsection{Transformer-Based NAG Model} 
	Given an input, \ie the vectorial image features in image captioning or the sentence tokens in machine translation, 
	NAG generates a sentence about the input in a non-autoregressive manner. 
	The architecture of our NAG model is based on the well-known Transformer network,  
	which composes of an encoder and decoder, as is shown in Figure~\ref{fig:framework}. 
	We only make the minimum necessary modifications on Transformer so as to maximize the decoding speed. 
	
	\paragraph*{Inputs}
	For machine translation task, the input are simply the tokens in the source sentence, which are converted to vectors of dimension $d$ with learned embeddings. 
	For image captioning task, the input image is first represented as a set of vectorial features that are either grid features extracted from a pre-trained CNN \cite{xu2015show} 
	or region features extracted from a pre-trained object detector \cite{anderson2017bottom}. 
	Each vector corresponds to a spatial position in the feature map when using grid features, 
	and corresponds to an object in the image when using region features. 
	We then feed the visual features through an input embedding layer to reduce the channel dimension to $d$. This embedding layer consists of a fully-connected layer followed by a ReLU and a
	dropout layer. 
	
	\paragraph*{Encoder} 
	The encoder of NAG is basically the same as the Transformer encoder, which takes the vectors as inputs and generates the context representation. 
	
	\paragraph*{Decoder} 
	Since the sequential dependency is removed in a non-autoregressive decoder, previous works often introduce additional components \eg well-designed decoder architecture \cite{pmlr-v80-kaiser18a,libovicky-helcl-2018-end} and decoder inputs \cite{guo2019non} \etc, which adds on extra inference time. 
	Different from these methods, we choose a design that simplifies the decoder to the most degree but proves to work well in our experiment. 
	We keep the decoder architecture almost the same as the Transformer decoder, 
	and simply use a sequence of sinusoidal positional encodings \cite{vaswani2017attention} as the decoder inputs, 
	each of which represents a position in the target sequence. 
	The sequence length is either fixed to $N$ (see Sec. \ref{sec:fixed}) or predicted by a target length predictor (see Sec. \ref{sec:predictor}). 
	We remove the causal attention mask from the self-attention layers of the decoder, 
	allowing each position to attend over all positions in the decoder.

	\subsection{NAG as a MARL System}
	To address the decoding inconsistency problem caused by word-level XE loss, 
	we propose to formulate NAG model as a fully cooperative Multi-Agent Reinforcement Learning (MARL) system, which optimizes sentence-level rewards and thus improves the decoding consistency. 
	We now formally cast NAG in MARL terminology.
	
	\paragraph*{Agent} Each word position in the target sequence is viewed as an agent that interacts with
	a common \textit{environment} (context from the encoder output) and other agents. There are $N$ agents in total, 
	identified by $a \in A \equiv\{a_1, a_2, \ldots, a_N\} $. 
	We denote joint quantities over all agents in bold, \eg $\mathbf{u}$, $\boldsymbol{\pi}$. 
	
	\paragraph*{State} The hidden states in NAG decoder layers naturally represent the states of the agents, 
	which are updated in each decoder layer. 
	The agents observe the environment through the inter-attention layer where they attend to the context.  
	The agents communicate with each other through the self-attention layer where the messages are passed between every two agents.  
	After $L$ rounds of observation and communication, the final state of each agent is denoted as $s_a$. 
	
	\paragraph*{Action} After obtaining $s_a$, each agent simultaneously chooses an action $u_a \in U$, 
	which is a word from the whole vocabulary $U$.
	The actions of all agents form a joint action $\mathbf{u} \in \mathbf{U} \equiv U^{N}$. %
	To transform the joint action into a sentence, we truncate the word sequence at the first end-of-sentence token ($<$eos$>$) when the number of agents is fixed, 
	and directly decode a sentence when the number of agents is decided by a target length predictor.  
	
	\paragraph*{Policy} The parameters of the network, $\theta$, define a stochastic policy $\pi_a$ for each agent, 
	from which the action is sampled, \ie $u_a \sim \pi_a = \textit{softmax}(s_a)$. 
	We speed up learning and reduce model complexity by sharing parameters among agents. 
	
	\paragraph*{Reward} 
	After all agents take their actions (words),  
	they receive a shared \textit{team reward} $R(\mathbf{u})$.  
	The reward is computed with an evaluation metric (\eg CIDEr) 
	by comparing the generated sentence to corresponding ground-truth sequences.

	Compared to typical MARL applications, NAG has two notable features. 
	First, NAG has a much larger action space (\ie the whole vocabulary, which is near 10,000 words). 
	Second, NAG has extremely shorter episodes (\ie the episode length is 1).  
	Actually, agents in NAG perform a one-step Markov Decision Process (MDP) since all words are generated in one-pass. 
	
	\begin{figure*}[ht]
		\begin{subfigure}{.5\textwidth}
			\centering
			\includegraphics[width=.8\linewidth]{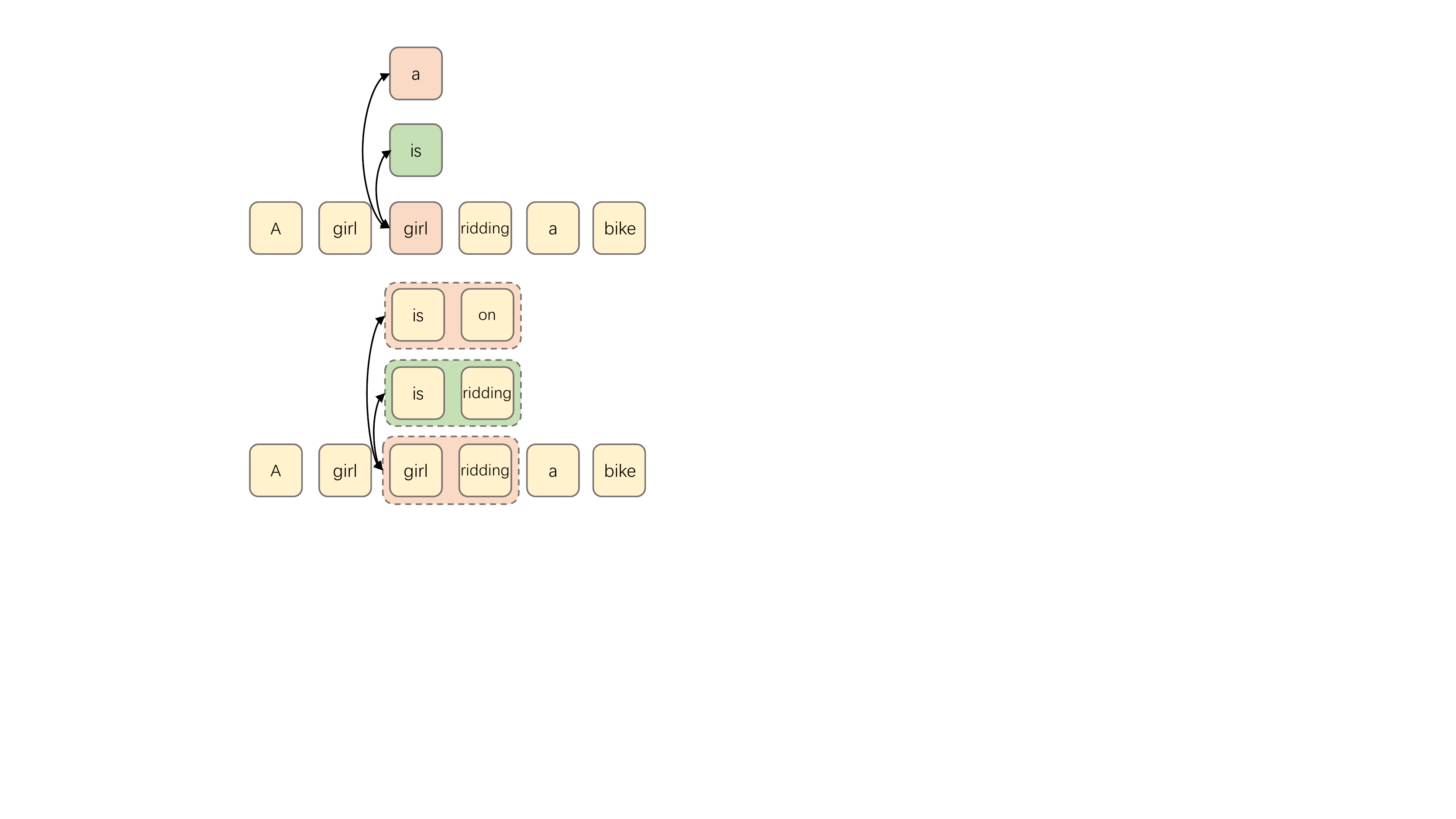}  
			\caption{Counterfactual replacements of an individual agent.}
			\label{fig:replace1}
		\end{subfigure}
		\begin{subfigure}{.5\textwidth}
			\centering
			\includegraphics[width=.8\linewidth]{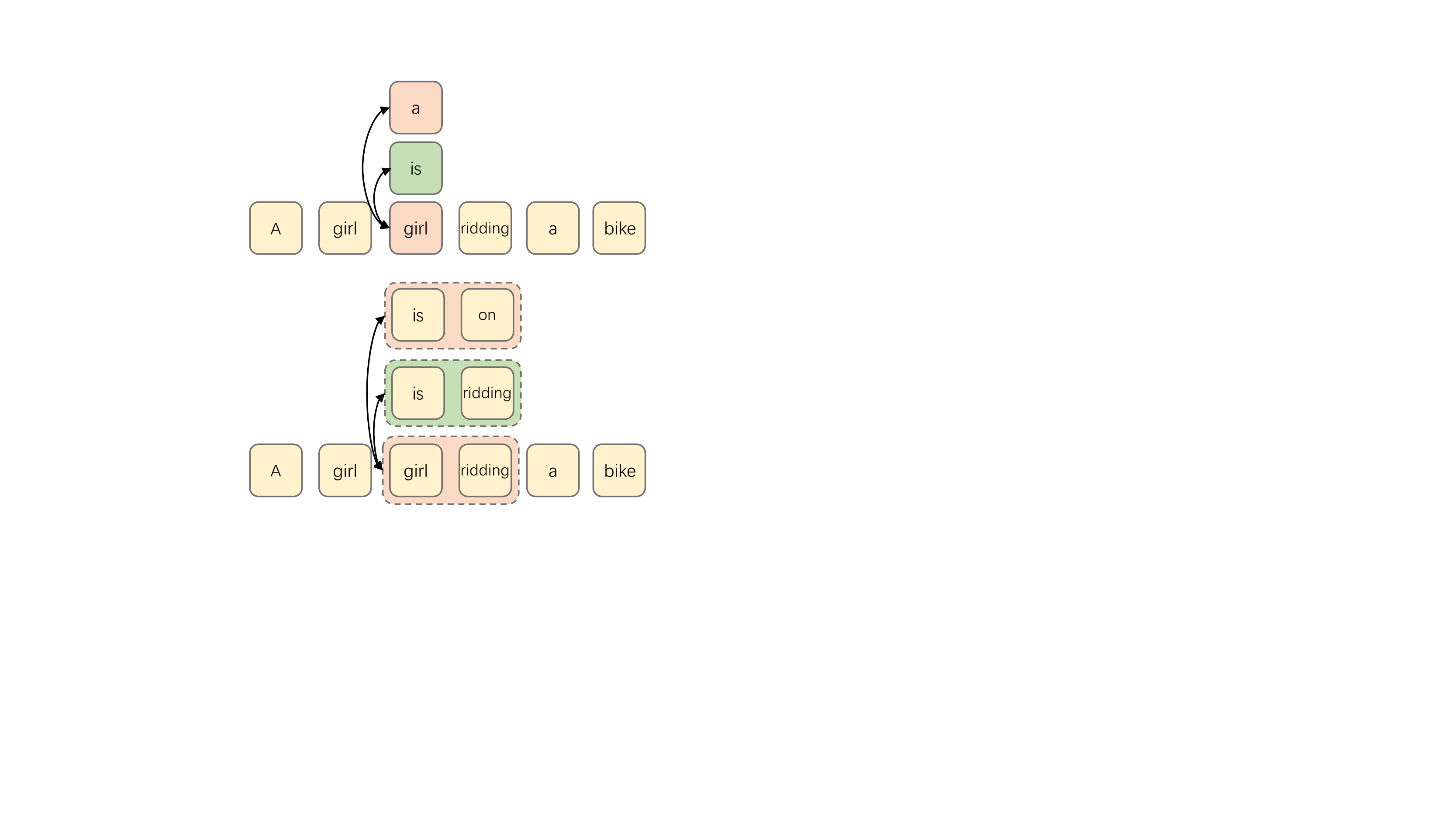}  
			\caption{Counterfactual replacements of a compositional agent.}
			\label{fig:replace2}
		\end{subfigure}
		\caption{Illustration of counterfactual replacements. We replace the action of an individual agent or a compositional agent while keeping the other agents' actions fixed.}
		\label{fig:replace}
	\end{figure*}
	
	\subsection{Multi-Agent Policy Gradient} 
	The goal of multi-agent learning is to maximize the expected team reward: 
	\begin{equation}
		\mathcal{L}(\theta)=-\mathbb{E}_{\boldsymbol{\pi}} \left[ R(\mathbf{u}) \right]. 
	\end{equation} 
	With the policy gradient theorem, the expected gradient of the agents can be computed as follows: 
	\begin{equation}
		\nabla_{\theta} \mathcal{L}(\theta)=-\mathbb{E}_{\boldsymbol{\pi}} \left[ \sum_a R(\mathbf{u})  \nabla_{\theta} \log \pi_a \left(u_a|s_a; \theta \right)\right]. 
	\end{equation} 
	Particularly, using the REINFORCE \cite{Williams1992Simple} algorithm, 
	the above equation can be approximated using a single sampling $\mathbf{u} \sim \boldsymbol{\pi} $ from the agents: 
	\begin{equation}
		\nabla_{\theta} \mathcal{L}(\theta) \approx  \sum_a R(\mathbf{u})  \nabla_{\theta} \log \pi_a \left(u_a|s_a; \theta \right) . 
	\end{equation}
	
	However, such a gradient estimate suffers from high variance, 
	which leads to unstable and slow learning of the optimal policy. 
	To reduce the variance, a reference reward or \textit{baseline} $b$ can be subtracted from the reward: 
	\begin{equation}
		\nabla_{\theta} \mathcal{L}(\theta) \approx  \sum_a (R(\mathbf{u}) - b)  \nabla_{\theta} \log \pi_a \left(u_a|s_a; \theta \right) . 
		\label{eqn:pg}
	\end{equation}
	The baseline still leads to an unbiased estimate, and importantly, it results in lower variance of the 
	gradient estimate \cite{sutton1998reinforcement}. 
	The baseline can be an arbitrary function, as long as it does not depend on the action $u_a$.

	\subsection{Individual Counterfactual Baseline}
	
	The above approach, however, fails to address a key multi-agent credit assignment problem. 
	That is, because each agent receives the same team reward, 
	it is unclear how a specific agent's action contributes to that team reward. 
	The consequences of this problem are inefficient multi-agent learning and decoding inconsistency.  
	For example, as illustrated in Figure \ref{fig:replace1}, suppose there is a generated sentence (joint action), ``a girl \underline{girl} riding a bike", and it gets a relatively high reward, 
	then the word ``girl" taken by the third agent is likely to be pushed up because it receives a positive reward, 
	which, however, should actually be suppressed and replaced with ``is". 
	
	To address this problem, we decide to compute a separate advantage function $A_a(s_a, \mathbf{u})$ for each agent. 
	It is computed by 
	subtracting an agent-specific \textit{counterfactual baseline} $B_a( s_a, \mathbf{u}_{-a} )$ from the common team reward, \ie:
	\begin{equation}
		A_a(s_a, \mathbf{u}) = R(\mathbf{u}) - B_a( s_a, \mathbf{u}_{-a} ), 
	\end{equation}
	where $\mathbf{u}_{-a}$ denotes the joint action of all the agents other than agent $a$. 
	$A_a(s_a, \mathbf{u})$ measures the increase (or decrease) in expected return of a joint action $\mathbf{u}$ due to agent $a$ having chosen action $u_a$ under state $s_a$. 
	The gradient in Eqn. (\ref{eqn:pg}) then becomes:
	\begin{equation}
		\nabla_{\theta} \mathcal{L}(\theta) \approx  \sum_a A_a(s_a, \mathbf{u})  \nabla_{\theta} \log \pi_a \left(u_a|s_a; \theta \right) . 
		\label{eqn:pg-cf}
	\end{equation}
	Since $B_a( s_a, \mathbf{u}_{-a} )$ does not depend on the action of agent $a$,  
	as described above, it will not change the expected gradient. 
	
	Formally, the counterfactual baseline $B_a$ is calculated by marginalizing the rewards when agent $a$ traverses all possible actions while keeping the other agents' actions $\mathbf{u}_{-a}$ fixed:
	
	\begin{equation}
		B_a( s_a, \mathbf{u}_{-a} ) = \mathbb{E}_{u^\prime_a \sim \pi_a} \left[ R([\mathbf{u}_{-a}, u^\prime_a]) \right].
		\label{eqn:expect}
	\end{equation}
	
	The key insight of using this counterfactual baseline for NAG is that: 
	given a sampled sequence/joint-action, if we replace the chosen word/action of a target position/agent with all possible words/actions and see how such counterfactual replacements affect the resulting reward, then the expected reward can act as a baseline to tell the actual influence of the chosen word/action. 
	As a result, for each agent, only actions that outperform its counterfactual baseline would be pushed up, 
	and inferior actions would be suppressed. 
	
	Because the action space of each agent is quite large, 
	we approximate the expectation computation in the above equation by only considering $k$ actions with the highest probability: 
	\begin{equation}
		\begin{aligned}
			\pi_a^\prime \left(u_a|s_a; \theta \right)& = \frac{ \pi_a \left(u_a|s_a; \theta \right) }{ \sum_{u^\prime_a \in \mathcal{T}_{a}} \pi_a \left(u^\prime_a|s_a; \theta \right) }, \\
			B_a( s_a, \mathbf{u}_{-a} ) &\approx \sum_{u^\prime_a \in \mathcal{T}_{a}} \pi_a^\prime \left(u^\prime_a|s_a; \theta \right)  R([\mathbf{u}_{-a}, u^\prime_a]), 
			\label{eqn:topk}
		\end{aligned}
	\end{equation}
	where $\pi^\prime \left(u_a|s_a; \theta \right)$ is the re-normalized probability for action $u_a$, and $\mathcal{T}_{a}$ is the set of words with top-$k$ probabilities in $\pi_a$. 
	Experimentally, we found this approximation to be quite accurate even with a relatively small $k$ 
	because the top-ranking words often have dominating probabilities. 
	
	Thanks to the one-step Markov Decision Process (MDP) nature of our NAG model, 
	the counterfactual replacements could be effortlessly made 
	by simply choosing new words from $ \pi_a \left(u_a|s_a; \theta \right)$, without the need for time-consuming Monte-Carlo rollouts as in common multi-step MDP problems.

	\subsection{Compositional Counterfactual Baseline}
	In the previous section we analyze an agent's contribution by counterfactual replacements of that single agent's action while keeping the other agents' actions fixed.  
	To take into account the joint influence of two neighboring agents, we introduce \textit{compositional agent} (CA). 
	Specifically, we treat two neighboring agents as a single compositional agent, and see how its \textit{compositional action} affects the team reward by the aforementioned counterfactual replacements of its action.  
	Take Figure \ref{fig:replace2} for example, for the generated sentence, ``a girl \underline{girl riding} a bike", we can consider the third and fourth agent as a compositional agent, and perform counterfactual replacements on its compositional action, \eg replacing ``girl riding" with ``is on" or ``is riding". 
	
	We denote the compositional agent of agent $a_i$ and its subsequent agent $a_{i+1}$ as $\tilde{a}_i=[a_i; a_{i+1}] $. 
	Similarly to Eqn. (\ref{eqn:expect}), for compositional agent $\tilde{a}_i$, its compositional counterfactual baseline $\widetilde{B}_{\tilde{a}_i}$ is calculated by marginalizing the rewards when agent $\tilde{a}_i$ traverses all possible compositional actions $u^\prime_{\tilde{a}_i} = [u_{a_i}; u_{a_{i+1}}]$ while keeping the other agents' actions $\mathbf{u}_{-{\tilde{a}_i}}$ fixed:
	
	\begin{equation}
		\widetilde{B}_{\tilde{a}_i}( s_{\tilde{a}_i}, \mathbf{u}_{-{\tilde{a}_i}} ) = \mathbb{E}_{u^\prime_{\tilde{a}_i} \sim \pi_{\tilde{a}_i}} \left[ R([\mathbf{u}_{-{\tilde{a}_i}}, u^\prime_{\tilde{a}_i}]) \right].
		\label{eqn:expect1}
	\end{equation}
	
	Also similar to Eqn. (\ref{eqn:topk}), we approximate the expectation computation in the above equation by only considering $k$ compositional actions with the highest probability: 
	\begin{equation}
		\begin{aligned}
			\pi_{\tilde{a}_i} \left(u_{\tilde{a}_i}|s_{\tilde{a}_i}; \theta \right) &= \pi_{{a}_i} \left(u_{{a}_i}|s_{{a}_i}; \theta \right) \cdot \pi_{{a}_{i+1}} \left(u_{{a}_{i+1}}|s_{{a}_{i+1}}; \theta \right), \\
			\pi_{\tilde{a}_i}^\prime \left(u_{\tilde{a}_i}|s_{\tilde{a}_i}; \theta \right)& = \frac{ \pi_{\tilde{a}_i} \left(u_{\tilde{a}_i}|s_{\tilde{a}_i}; \theta \right) }{ \sum_{u^\prime_{\tilde{a}_i} \in \mathcal{T}_{{\tilde{a}_i}}} \pi_{\tilde{a}_i} \left(u^\prime_{\tilde{a}_i}|s_{\tilde{a}_i}; \theta \right) }, \\
			\widetilde{B}_{\tilde{a}_i}( s_{\tilde{a}_i}, \mathbf{u}_{-{\tilde{a}_i}} ) &\approx \sum_{u^\prime_{\tilde{a}_i} \in \mathcal{T}_{{\tilde{a}_i}}} \pi_{\tilde{a}_i}^\prime \left(u^\prime_{\tilde{a}_i}|s_{\tilde{a}_i}; \theta \right)  R([\mathbf{u}_{-{\tilde{a}_i}}, u^\prime_{\tilde{a}_i}]), 
			\label{eqn:topk1}
		\end{aligned}
	\end{equation}
	where $\pi_{\tilde{a}_i} \left(u_{\tilde{a}_i}|s_{\tilde{a}_i}; \theta \right)$ and $\pi_{\tilde{a}_i}^\prime \left(u_{\tilde{a}_i}|s_{\tilde{a}_i}; \theta \right)$ are the joint probability and the re-normalized probability for compositional action $u_{\tilde{a}_i}$, and 
	$\mathcal{T}_{{\tilde{a}_i}}$ is the set of words with top-$k$ probabilities in $\pi_{\tilde{a}_i}^\prime$. 
	
	Since an agent $a_i$ has two neighboring agents, \ie its preceding agent $a_{i-1}$ and subsequent agent $a_{i+1}$, it involves in two compositional agents, $\tilde{a}_{i-1}=[a_{i-1}; a_{i}] $ and $\tilde{a}_i=[a_i; a_{i+1}] $. 
	Therefore, for agent $a_i$, its involved compositional counterfactual baseline is the average of $\widetilde{B}_{\tilde{a}_{i-1}}$ and $\widetilde{B}_{\tilde{a}_i}$: 
	\begin{equation}
		\begin{aligned}
			\widetilde{B}_{a_i} = \frac{\widetilde{B}_{\tilde{a}_{i-1}} + \widetilde{B}_{\tilde{a}_i}}{2}.
			\label{eqn:2b}
		\end{aligned}
	\end{equation}
	
	The final counterfactual baseline for agent $a_i$ is the weighted-sum of two parts: the individual baseline $B_{a_i}$ in Eqn. (\ref{eqn:topk}) and the compositional baseline $\widetilde{B}_{a_i}$ in Eqn. (\ref{eqn:2b}):
	\begin{equation}
		\widehat{B}_{a_i} = (1-\lambda) B_{a_i} + \lambda \widetilde{B}_{a_i}, 
		\label{eqn:final_baseline}
	\end{equation}
	where $\lambda$ is a hyper-parameter to balance the two terms. 
	We empirically found set $\lambda$ to $0.5$ performs well. 
	We only consider the compositional agent of two neighboring agents, since in our preliminary experiment we did not observe further improvement when using more adjacent agents. 
	
	Finally, by considering compositional agent (CA), the gradient in Eqn. (\ref{eqn:pg}) becomes:
	\begin{equation}
		\begin{aligned}
			\nabla_{\theta} \mathcal{L}(\theta)  &\approx  \sum_a \left(R(\mathbf{u}) - \widehat{B}_a\right)  \nabla_{\theta} \log \pi_a \left(u_a|s_a; \theta \right) . 
		\end{aligned}
		\label{eqn:pg-cf1}
	\end{equation}
	
	The individual baseline measures the expected effect of an agent to the team reward by considering that individual agent's actions, 
	while the compositional baseline further considers how that agent's actions interact with its neighbor agents' actions. 
	The individual baseline and compositional baseline are complementary in that 
	the combination of they can more precisely measure the expected reward of an agent 
	and hence more accurately measure how much gain the agent brings by taking the sampled action. 
	
	\begin{table*}[htbp]
		\centering
		\caption{Generation Quality, Latency, and Speedup on MSCOCO Image Captioning Dataset. 
			``$\dag$" Indicates the Model is Based on LSTM Architecture While the others are Based on Transformer. 
			AIC is Our Implementation of the Transformer-Based Autoregressive Model, Which Has the Same Structure as NAIC Models and is Used as the Teacher Model for Knowledge Distillation (KD). 
			``/" Denotes That the Results are not Reported or Not Directly Comparable. 
			``bw" denotes the Beam Width Used for Beam Search. 
			Latency is the Time to Decode a Single Image without Minibatching, 
			Averaged Over the Whole Test Split, and is Tested on a GeForce GTX 1080 Ti GPU.
			The Speedup Values of the Compared Models are from the Corresponding Papers. 
		}
		\label{tab:main}%
		\begin{tabular}{l|cccccc|cc}
			\toprule
			Models & \multicolumn{1}{c}{BLEU-1} & \multicolumn{1}{c}{BLEU-4} & \multicolumn{1}{c}{METEOR} & \multicolumn{1}{c}{ROUGE} & \multicolumn{1}{c}{SPICE} & \multicolumn{1}{c}{CIDEr} & \multicolumn{1}{|c}{Latency} & \multicolumn{1}{c}{Speedup} \\
			\midrule
			\midrule
			\multicolumn{9}{l}{\textbf{Autoregressive models}} \\
			\midrule
			NIC-v2$^\dag$ \cite{vinyals2017show} & / & 32.1  & 25.7  & / & / & 99.8  & / & / \\
			Up-Down$^\dag$ \cite{anderson2017bottom} & 79.8  & 36.3  & 27.7  & 56.9  & 21.4  & 120.1 & / & / \\
			CAVP$^\dag$ \cite{zha2019context}  & / & 38.6 & 28.3 &58.5 &21.6 & 126.3 & / & / \\
			AoANet$^\dag$ \cite{huang2019attention}& 80.2 & 38.9 & 29.2 & 58.8 & 22.4 & 129.8 & / & / \\
			ETA \cite{li2019entangled} & 81.5 & 39.3 & 28.8 & 58.9 & 22.7 & 126.6  & / & / \\
			ORT \cite{herdade2019image}   & 80.5  & 38.6  & 28.7  & 58.4  & 22.6  & 128.3 & / & / \\
			NG-SAN \cite{guo2020normalized} & / & 39.9 &29.3 &59.2 &23.3 & 132.1 & / & / \\
			X-Transformer \cite{pan2020x}& 80.9 & 39.7 & 29.5 &59.1 &23.4 & 132.8 & / & / \\
			AIC ($\text{bw}=1$) &  79.8&	38.4&	29.0 &	58.7	& 22.8&	126.6  &   134ms    &  1.66$\times$ \\
			AIC ($\text{bw}=3$) &  80.3&	38.9&	29.1 &	58.9	& 22.9&	128.8  &   222ms    &  1.00$\times$ \\
			\midrule 
			\multicolumn{9}{l}{\textbf{Non-autoregressive models}} \\
			\midrule
			MNIC \cite{gao2019masked} & 75.4  & 30.9  & 27.5  & 55.6  & 21.0    & 108.1 & / & 2.80$\times$ \\
			FNIC \cite{fei2019fast} 						  & /     & 36.2  & 27.1  & 55.3  &20.2     &115.7  & /   &  8.15$\times$ \\
			\midrule
			\multicolumn{9}{l}{\textbf{Non-autoregressive models (Ours)}} \\
			\midrule
			NAIC-base   &   60.7	& 15.9&	18.2&	45.9&	11.9	& 60.6   &    \multirow{7}[1]{*}{\bf16ms}  &  \multirow{7}[1]{*}{\bf13.90$\times$} \\
			\ \ \ \ + weight-init   &  62.3 &	17.1	 &19.0 &	46.8 &	12.6 &	64.6  &       &  \\
			\ \ \ \ + KD &  78.5&	35.3	&27.3 &	56.9	& 20.8&	115.5   &       &  \\
			\ \ \ \ + CMAL &  80.3 &	37.3&	28.1	&58.0 & 	21.8	& 124.0  &       &  \\
			\ \ \ \ + unlabel  &    80.5 &	38.0 &	28.3 &	58.2	& 22.0 &	125.5   &       &  \\
			\ \ \ \ + CA  &    80.6  & 	38.2  & 	28.4  & 	58.4  & 	22.1  & 	126.4  &   &   \\
			\ \ \ \ + postprocess  & \bf	80.7  &	\bf 38.4 &\bf	28.4 &\bf	58.5 &	\bf22.1 &\bf	126.9 &  & \\
			\bottomrule
		\end{tabular}%
	\end{table*}%

	\subsection{Unlabeled Data Augmented Knowledge Distillation} 
	Following previous works \cite{gu2017non}, we use sequence-level knowledge distillation (\textbf{KD}) \cite{kim2016sequence} during training, 
	where the sentences produced by a pre-trained autoregressive Transformer teacher model is considered as pseudo target sentences for training the non-autoregressive student model. 
	This strategy has been shown to be an effective way to alleviate the multimodality problem \cite{gu2017non}. 
	
	While previously only labeled data are used for training non-autoregressive models, 
	we propose to utilize freely available unlabeled inputs to boost the performance of sequence generation with the KD strategy.  
	Specifically, we use KD teacher model to produce target sentences for extra unlabeled data, which are used as pseudo paired data during XE training. 
	These unlabeled data can be seen as a data augmentation technique that helps to better transfer the teacher model's knowledge to the student model. 
	
	Before starting CMAL training, we first pre-train the NAG models with the XE loss (Eqn. (\ref{eqn:xe})), 
	during which we use both the labeled and unlabeled data and their corresponding \textit{pseudo} target sentences as training data. 
	Then during CMAL training (Eqn. (\ref{eqn:pg-cf}) and Eqn. (\ref{eqn:pg-cf1})), we use the \textit{real} labeled data from the original dataset.  
	There are two advantages of using real data instead of pseudo data for CMAL training: 
	first, the reward computation at training time is consistent with the evaluation metric computation at test time,  
	\ie the generated sentences are compared against the real target sentences;  
	second, unlike previous works on NAG, 
	the performance of our method will not be limited by that of the KD teacher model.

	\section{Experiments on Image Captioning}
	We first validate the proposed non-autoregressive sequence generation method on image captioning task. 
	We denote our Non-Autoregressive Image Captioning models as \textbf{NAIC}.
	
	\subsection{Experimental Settings} 
	\subsubsection{MSCOCO Dataset} MSCOCO \cite{chen2015microsoft} 
	is the most popular benchmark for image captioning. 
	We use the ``Karpathy" splits \cite{karpathy2015deep} that have been used extensively for 
	reporting results in prior works. This split contains 113,287
	training images with 5 captions each, and 5,000 images for validation and test splits, respectively. 
	The vocabulary size is 9,487 words. 
	We use the officially released MSCOCO unlabeled images as unlabeled data. 
	To be consistent with previous works, we pre-extract image features for all the images following \cite{anderson2017bottom}. 
	We use standard automatic evaluation metrics to evaluate the quality of captions, including BLEU-1/4 \cite{Papineni2002BLEU}, METEOR \cite{Denkowski2014Meteor}, ROUGE \cite{lin2004rouge}, SPICE \cite{Anderson2016SPICE}, and CIDEr \cite{Vedantam2015CIDEr}, 
	denoted as B1/4, M, R, S, and C, respectively.

	\subsubsection{Implementation Details}
	\label{sec:fixed}
	We train an autoregressive image captioning model as the teacher model, namely AIC. 
	Both NAIC and the AIC models closely follow the same model hyper-parameters as Transformer-\textit{Base} \cite{vaswani2017attention}.  
	Specifically, the number of stacked blocks $L$ is 6. 
	The AIC model is trained first with XE loss and then with SCST \cite{Rennie2016Self}. 
	Beam search with a beam width of 3 is used during decoding of AIC model. 
	Our best NAIC model is trained according to the following process. 
	We use a fixed number of $N=16$ agents because most of the captions are not longer than this length. 
	We first initialize the weights of NAIC model with the pre-trained AIC teacher model. 
	We then pre-train NAIC model with XE loss for 30 epochs. 
	During this stage, we use a warm-up learning rate of 
	$min( t\times 10^{-4}; 3\times 10^{-4})$, where $t$ is the current epoch number that starts at 1. 
	After 6 epochs, the learning rate is decayed by 0.5 every 3 epochs. 
	After that, we run CMAL training to optimize the CIDEr metric for about 70 epochs. 
	To enhance training efficiency, we sample 5 sentences for each input. 
	At this training stage, we use an initial learning rate of $7.5\times 10^{-5}$ and decay it by 0.8 every 10 epochs.
	Both training stages use Adam \cite{kingma2014adam} optimizer with a batch size of 50. 
	By default, we use $k=2$ top-ranking words in CMAL, and use $100,000$ unlabeled images for training.

	\subsection{Main Results} 
	We first compare the performance of our methods against other non-autoregressive models and state-of-the-art autoregressive models. 
	Among the autoregressive models, ETA \cite{li2019entangled}, ORT \cite{herdade2019image}, X-Transformer \cite{pan2020x}, NG-SAN \cite{guo2020normalized}, MNIC \cite{gao2019masked}, FNIC\cite{fei2019fast}, and AIC are based on similar Transformer architecture as ours, while others are based on LSTM \cite{hochreiter1997long}. 
	MNIC \cite{gao2019masked} and FNIC\cite{fei2019fast} are two published non-autoregressive image captioning models. 
	MNIC adopts an iterative refinement strategy while FNIC uses an RNN to order words detected in the image. 
	
	As shown in Table~\ref{tab:main}, our best model (the last row) 
	achieves significant improvements over the previous non-autoregressive models across all metrics,
	strikingly narrowing their performance gap between AIC from $13.1$ CIDEr points down to only $1.9$ CIDEr points. 
	Furthermore, we achieve comparable performance with state-of-the-art autoregressive models.  
	Comparing speedups, our method obtains a significant speedup of more than a factor of 10 over the 
	autoregressive counterpart, with latency\footnote{The time for image feature extraction is not included in latency. } reduced to about $16ms$. 
	In Table~\ref{tab:server} we further compare our NAIC model with state-of-the-art autoregressive models by submitting our single-model results to online MSCOCO evaluation server. C5 and c40 denote evaluating the models with 5 and 40 reference captions, respectively. 
	The results show that our method achieves very competitive results compared to state-of-the-art autoregressive models and even surpasses some of them.

	\begin{table*}[tbp]
		\centering
		\caption{Comparision with State-Of-the-Art, \textit{Single-Model} Autoregressive Methods on the Online MSCOCO Image Captioning Test Server. 
		}
		\label{tab:server}%
		\begin{tabular}{@{\extracolsep{3pt}}@{\kern\tabcolsep}lrrrrrrrrrrrrrr}
			\toprule
			\multicolumn{1}{c}{\multirow{2}[4]{*}{Model}}  & \multicolumn{2}{c}{BLEU-1} & \multicolumn{2}{c}{BLEU-2} & \multicolumn{2}{c}{BLEU-3} & \multicolumn{2}{c}{BLEU-4} & \multicolumn{2}{c}{METEOR} & \multicolumn{2}{c}{ROUGE-L} & \multicolumn{2}{c}{CIDEr-D} \\
			\cmidrule{2-3}  \cmidrule{4-5} \cmidrule{6-7} \cmidrule{8-9} \cmidrule{10-11} \cmidrule{12-13} \cmidrule{14-15} %
			& \multicolumn{1}{c}{c5} & \multicolumn{1}{c}{c40} & \multicolumn{1}{c}{c5} & \multicolumn{1}{c}{c40} & \multicolumn{1}{c}{c5} & \multicolumn{1}{c}{c40} & \multicolumn{1}{c}{c5} & \multicolumn{1}{c}{c40} & \multicolumn{1}{c}{c5} & \multicolumn{1}{c}{c40} & \multicolumn{1}{c}{c5} & \multicolumn{1}{c}{c40} & \multicolumn{1}{c}{c5} & \multicolumn{1}{c}{c40} \\
			\midrule
			Up-Down$^*$ \cite{anderson2017bottom} & 80.2  & 95.2  & 64.1  & 88.8  & 49.1  & 79.4  & 36.9  & 68.5  & 27.6  & 36.7  & 57.1  & 72.4  & 117.9  & 120.5  \\
			CAVP \cite{zha2019context}  & 80.1  & 94.9  & 64.7  & 88.8  & 50.0  & 79.7  & 37.9 & 69.0  & 28.1  & 37.0  & 58.2  & 73.1  & 121.6  & 123.8  \\
			SGAE \cite{yang2019auto} & 80.6  & 95.0  & 65.0  & 88.9  & 50.1  & 79.6  & 37.8  & 68.7  & 28.1  & 37.0  & 58.2  & 73.1  & 122.7  & 125.5  \\
			ETA \cite{li2019entangled} & 81.2& 95.0 &65.5 &89.0& 50.9& 80.4& 38.9& 70.2& 28.6 &38.0& 58.6 &73.9& 122.1& 124.4 \\
			NG-SAN \cite{guo2020normalized}  & 80.8 & 95.0 &65.4& 89.3& 50.8& 80.6& 38.8& 70.2 &29.0 &38.4 &58.7& 74.0& 126.3 &128.6 \\
			\midrule
			Ours & 80.3 & 94.4 & 64.5 & 87.8 & 49.6 & 78.2 & 37.5 & 67.4 & 28.1 & 36.8 & 58.0 & 72.8 & 121.1 & 122.6 \\
			\bottomrule
		\end{tabular}%
	\end{table*}%

	\subsection{The Contribution of Each Component}
	We conduct an extensive ablation study with the proposed NAIC model. 
	The results are shown in the bottom of Table~\ref{tab:main}. 
	\textit{NAIC-base} is the naive NAIC model trained from scratch using XE loss (Eqn. (\ref{eqn:xe})), 
	\textit{weight-init} denotes initializing the weights of NAIC with AIC model, 
	\textit{KD} represents using the sentence-level knowledge distillation with AIC as the teacher model, 
	\textit{CMAL} denotes further applying our proposed CMAL algorithm with individual counterfactual baseline (Eqn.~(\ref{eqn:pg-cf})) for CIDEr optimization, 
	\textit{unlabel} means using additional 100,000 unlabeled data during XE training,  
	\textit{CA} indicates applying compositional counterfactual baseline with Eqn.~(\ref{eqn:pg-cf1}),  
	and \textit{postprocess} is a simple postprocessing step whcih removes the consecutive identical tokens following\cite{lee2018deterministic} .

	From Table~\ref{tab:main}, we have the following observations. 
	(1) Initializing NAIC model's weights with the AIC teacher can consistently improve the performance.  
	(2) NAIC-base performs extremely poorly compared to AIC. 
	(3) We see that training on the distillation data during XE training improves the CIDEr score to 115.5. 
	However, there still remains a large performance gap between this model and the AIC teacher. 
	(4) Applying our CMAL training on top of the above XE trained model 
	significantly improves the performance by $8.5$ CIDEr points.  
	(5) Augmenting knowledge distillation data with extra unlabeled data during XE training further boosts the performance by $1.5$ CIDEr points. 
	(6) Considering the influence of compositional agents (CA) on CMAL's counterfactual baseline further increases the CIDEr score by $0.9$ points.
	(7) Removing the consecutive identical tokens (postprocess) slightly improve the CIDEr score by $0.59$ points, resulting in a final CIDEr score of $126.9$, 
	which is only $1.9$ points behind the AIC teacher. 
	
	\begin{table}[tp]
		\centering
		\caption{Comparison of Using Various Baselines $b$ (in Eqn. (\ref{eqn:pg}) on MSCOCO Image Captioning Dataset.  
			XE: The Performance After Pre-Training With Cross-Entropy Loss. 
			None: Not Using a Baseline. 
			MA: Moving Average Baseline. 
			SC: Self-Critical Baseline. 
			CF: The Proposed Counterfactual Baseline.
		}
		\label{tab:baseline}%
		\begin{tabular}{lrrrrrr}
			\toprule
			\multicolumn{1}{l}{Baseline $b$}  & \multicolumn{1}{c}{B1} & \multicolumn{1}{c}{B4} & \multicolumn{1}{c}{M} & \multicolumn{1}{c}{R} & \multicolumn{1}{c}{S} & \multicolumn{1}{c}{C} \\
			\midrule
			\midrule
			\multicolumn{1}{l}{\textbf{w/o weight-init:}} \\
			\ \ \ \ XE & 77.7 & 34.8 &	26.9 &	56.3 &	20.3 &	113.9 \\
			\ \ \ \ None & 65.6& 	19.4 &	22.7 &	48.9 &	15.8 &	91.4 \\
			\ \ \ \ MA & 75.6 &	28.7 &	24.4 &	53.6 &	17.9 &	103.3  \\
			\ \ \ \ SC & 79.0 	&34.6 &	26.9 &	56.2 &	20.6 &	118.1 \\
			\ \ \ \ CF & \bf79.9 &	\bf36.5 &	\bf27.7 &	\bf57.4 &	\bf21.4 &	\bf122.1 \\
			\midrule
			\multicolumn{1}{l}{\textbf{w/ weight-init:}} \\
			\ \ \ \ XE & 78.5 &	35.3 &	27.3 &	56.9 &	20.8 &	115.5 \\
			\ \ \ \ None & 78.6 &	33.7 &	26.5 &	56.1 &	20.2& 	115.2 \\
			\ \ \ \ MA & 79.0 &	34.1 	&26.6 	&56.3 	&20.2 &	116.1 \\
			\ \ \ \ SC & 79.6 	&36.5 &	27.6 &	57.4 	&21.4 &	121.2 \\
			\ \ \ \ CF & \bf80.3 &	\bf37.3 &	\bf28.1 	& \bf58.0 	&\bf21.8 &	\bf124.0 \\
			\bottomrule
		\end{tabular}%
	\end{table}%

	\subsection{Comparison of Various Reward Baselines $b$} 
	\label{sec:baseline}
	To evaluate the effectiveness of our counterfactual (CF) baseline, 
	we compare it with two widely-used baselines in policy gradient, 
	\ie Moving Average \cite{weaver2001optimal} (MA) and Self-Critical \cite{Rennie2016Self} (SC), and not using a baseline (None), \ie$b=0$. 
	Specially, CF baseline is the proposed individual counterfactual baseline (Eqn.~(\ref{eqn:pg-cf})). 
	MA baseline is the accumulated sum of the previous rewards with exponential decay. 
	SC baseline is the received reward when all agents directly take greedy actions.

	The results are shown in Table~\ref{tab:baseline}.
	We specially consider the case when not using weight-init  
	because it may not be possible to find an autoregressive model that has the same structure 
	as a novelly designed non-autoregressive model. 
	As we can see, our CF baseline consistently outperforms all the other compared methods. 
	Noteworthy that the performance gaps between our CF baseline and other baselines become larger when 
	trainings start from a low-performing model (\ie XE model w/o weight-init). 
	That is, our method is less sensitive to model initialization, 
	suggesting its ability to enable more robust and stable reinforcement learning.  
	None and MA severely degrade the performance compared to XE model when not using weight-init, 
	but they perform similar to XE model when using weight-init. 
	While SC considerably outperforms XE model, it is still inferior to CF. 
	The reason is that both MA and SC are agent-agnostic global baselines that 
	fail to address the multi-agent credit assignment problem. 
	Different from these methods, our CF baseline is agent-specific and can disentangle the individual contribution of each agent from the team reward so that the agents can be trained more efficiently.  
	\begin{figure*}[btp] 
		\centering
		\includegraphics[width=7in]{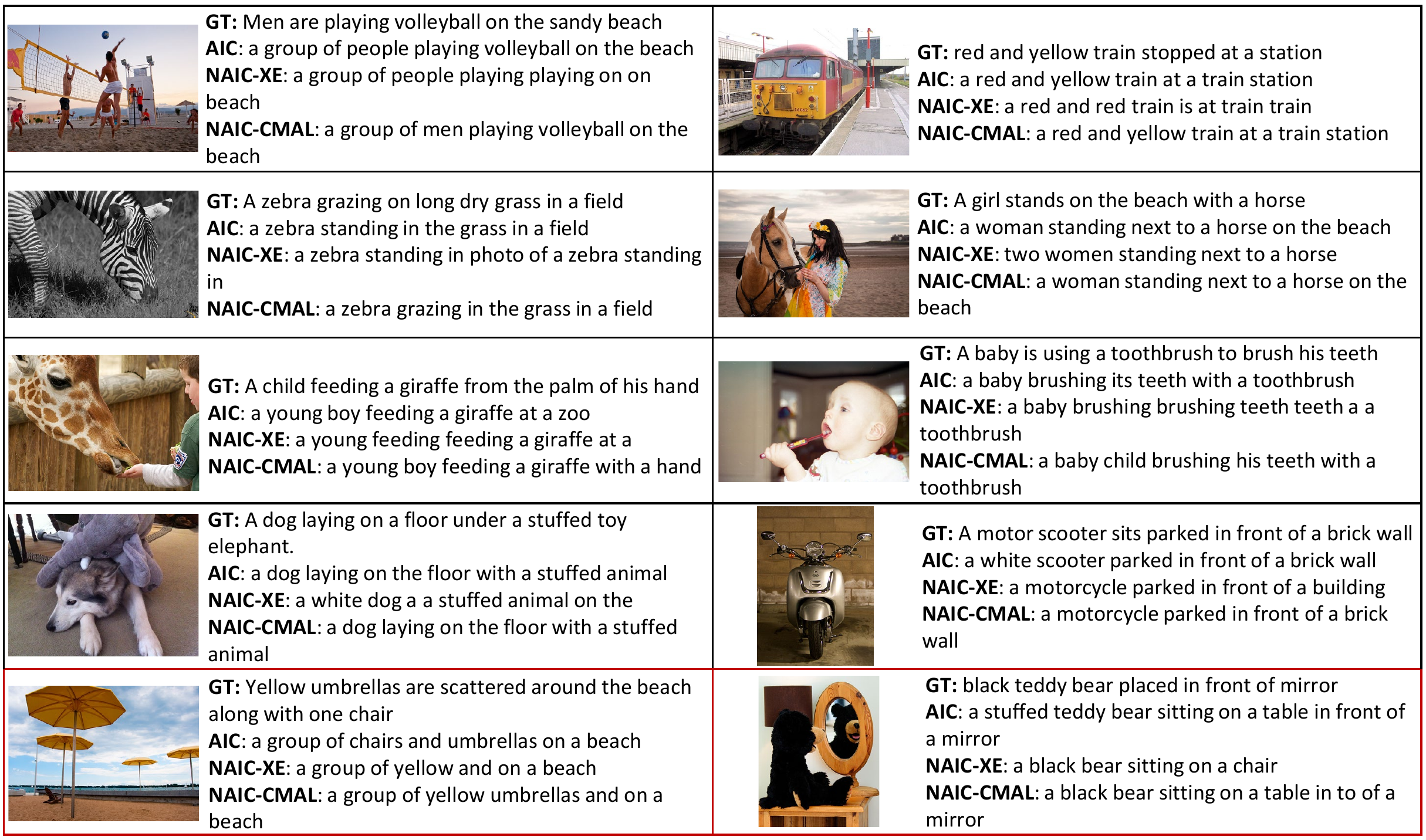}
		\caption{
			Examples of the generated image captions on MSCOCO image captioning dataset. 
			GT is a ground-truth caption. NAIC-XE and NAIC-CMAL are our NAIC model after XE and CMAL training, respectively. 
			Two failure cases are highlighted in red. 
		}
		\label{fig:example}
	\end{figure*}

	\begin{table}[tp]
		\centering
		\caption{Effect of Top-$k$ Size (in Eqn. (\ref{eqn:topk}) on MSCOCO Image Captioning Dataset. }
		\label{tab:topk}%
		\begin{tabular}{crrrrrr}
			\toprule
			\multicolumn{1}{c}{top-$k$} & \multicolumn{1}{c}{B1} & \multicolumn{1}{c}{B4} & \multicolumn{1}{c}{M} & \multicolumn{1}{c}{R} & \multicolumn{1}{c}{S} & \multicolumn{1}{c}{C} \\
			\midrule
			\midrule
			1 & 80.1 &	\bf37.4 &	28.0 &	57.9 &	21.7 &	123.7 \\
			2 & \bf80.3 &	 37.3 &	\bf28.1 	& \bf58.0 	&\bf21.8 &	\bf124.0 \\
			5 & 80.1 &	37.3 &	28.0 &	58.0 &	21.7 &	123.7 \\
			\bottomrule
		\end{tabular}%
	\end{table}%
	\begin{table}[tp]
		\centering
		\caption{Performance of Our Method as a Function of Training Metric on MSCOCO Image Captioning Dataset.}
		\label{tab:metric}%
		\begin{tabular}{lrrrrrr}
			\toprule
			\multirow{2}[0]{*}{\thead{Training\\Metric}} & \multicolumn{6}{c}{Evaluation Metric} \\
			\cmidrule{2-7}
			&\multicolumn{1}{c}{B1} & \multicolumn{1}{c}{B4} & \multicolumn{1}{c}{M} & \multicolumn{1}{c}{R} & \multicolumn{1}{c}{S} & \multicolumn{1}{c}{C} \\
			\midrule
			\midrule
			XE    &   78.5 &	35.3 &	27.3 &	56.9 	&20.8& 	115.5  \\
			BLEU-4  &   78.4 &	\bf38.3 &	27.6 &	58.0 &	21.1 &	118.4 \\
			METEOR & 77.9 &	35.5 &	\bf28.6 &	57.9 &	\bf22.2 &	119.5  \\
			ROUGE &  78.6 &	37.5 &	27.6 &	\bf58.6 &	20.9 &	118.9  \\
			CIDEr &  \bf80.3 &	37.3 &	28.1 &	58.0 &	21.8 &	\bf124.0  \\
			\bottomrule
		\end{tabular}%
	\end{table}%
	
	\subsection{Effect of Top-$k$ Size}
	As shown in Table~\ref{tab:topk}, the model is not sensitive to the choice of top-$k$ size.  
	Using a small $k$ of 2 could achieve fairly good performance.  
	This is because after pre-training NAIC model with cross-entropy loss, the top-ranking words often have dominating probabilities and thus 
	making it feasible to approximate the expectation over all words in the vocabulary with these $k$ top-ranking words.

	\begin{table}[tp]
		\centering
		\caption{The Results After XE and CMAL Training When Using Different Numbers of Unlabeled Images on MSCOCO Image Captioning Dataset.}
		\label{tab:unlabel}%
		\begin{tabular}{cc|rrrrrr}
			\toprule
			\#unlabel & stage & \multicolumn{1}{c}{B1} & \multicolumn{1}{c}{B4} & \multicolumn{1}{c}{M} & \multicolumn{1}{c}{R} & \multicolumn{1}{c}{S} & \multicolumn{1}{c}{C} \\
			\midrule
			\midrule
			\multirow{2}[0]{*}{0} & XE & 78.5 &	35.3 &	27.3 &	56.9 	&20.8& 	115.5  \\
			& CMAL &  80.3 	&37.3 &	28.1 &	58.0& 	21.8 &	124.0  \\
			\midrule
			\multirow{2}[0]{*}{50k} & XE & 78.8 	&36.2 &	27.6 &	57.2 &	21.1 	&118.1  \\
			& CMAL &   80.2 & 	37.6 & 	28.1 & 	58.1 & 	21.9 & 	124.8 \\
			\midrule
			\multirow{2}[0]{*}{100k} & XE &  79.0 &	36.2 &	27.7 &	57.3 &	21.2& 	118.3 \\
			& CMAL &   \bf80.5 &	\bf38.0 &	\bf28.3 &	\bf58.2 &	\bf22.0 &	\bf125.5 \\ 
			\bottomrule
		\end{tabular}%
	\end{table}%

	\subsection{Training on Different Metrics.} 
	In Table~\ref{tab:metric} we show the results of using different evaluation metrics as reward function in CMAL. 
	The SPICE metric is not experimented because it computation runs too slow. 
	As expected, optimizing for a given metric during training generally leads
	to the best performance on that same metric at test time. 
	Optimizing for CIDEr gives more balanced improvements on all the evaluation metrics. 
	We have experimented with optimizing various mixed metrics, but no noticeable improvement was observed.

	\subsection{Number of Unlabeled Images} 
	In Table~\ref{tab:unlabel}, we show the results after XE and CMAL training 
	when using 0, 50,000 and 100,000 unlabeled images respectively. 
	Generally, using more unlabeled images could lead to better performance.  
	XE training benefits more from the unlabeled images than CMAL training because 
	we directly use the unlabeled images during XE training while not using them for CMAL.

	{\renewcommand{\arraystretch}{1.7}%
		\begin{table*}[tp]
			\centering
			\caption{Two Examples Comparing Translations Produced by the autoregressive Transformer, the Non-autoregressive NAT-XE, and Our NAT-CMAL without Postprocessing.
				Repeated Words are Highlighted in Gray. }
			\label{tab:nmt_case}%
			\begin{tabular}{r|l}
				\hline
				Source: & NSA siphons data from Google and Yahoo - Snowden wants to help \\
				\hline
				Target: & NSA saugt Daten von Google und Yahoo ab - Snowden will helfen \\
				\hline
				Transformer:& NSA sickert Daten von Google und Yahoo - Snowden will helfen \\
				\hline
				NAT-XE: & NSA - phdaten von \colorbox{gray!30}{Google Google Yahoo Yahoo} SnowSnowden will helfen \\
				\hline
				NAT-CMAL: &NSA sionen Daten von Google und Yahoo - Snowden will helfen \\
				\hline
				\hline
				Source: & Helpers can include parents , grandparents , friends , organisations and , of course , the teachers and children themselves . \\
				\hline
				Target: & Helfer können Eltern , Großeltern , Freunde , Vereine und natürlich die Erzieher und Kinder selbst sein . \\
				\hline
				Transformer: & Helfer können Eltern , Großeltern , Freunde , Organisationen und natürlich die Lehrer und Kinder selbst sein . \\
				\hline
				\multirow{2}{*}{NAT-XE:} & Zu fer können Eltern GroßGroßeltern \colorbox{gray!30}{, , , Organisationen Organisationen natürlich natürlich} \\
				& \colorbox{gray!30}{Lehrer Lehrer Kinder Kinder} eingehören . \\
				\hline
				NAT-CMAL: & Zu Helfer können Eltern , Großeltern , Freunde , Organisationen und natürlich die Lehrer und Kinder selbst einschließen . \\
				\hline
			\end{tabular}%
		\end{table*}%
	}
	
	\subsection{Qualitative Analysis}
	We present some examples of generated image captions in Figure~\ref{fig:example}. 
	As can be seen, repeated words and incomplete content are most prevalent in the XE trained NAIC model, \eg ``playing playing on on", ``zebra standing ... zebra standing", and ``brushing brushing teeth teeth a a".  
	This proves that the word-level XE training often results in decoding inconsistency problem. 
	With our CMAL training, the sentences often become far more consistent and fluent. 
	We also show two failure cases in the last row, highlighted in red. 
	In the first case, our model outputs the unfluent sentence ``and on a beach", which might be due to its failure to recognize the ``chair". 
	In the second case, our model outputs ``in to of a mirror", which should be ``in front of a mirror". 
	
	\begin{table}[tp]
		\centering
		\caption{Statistics of MSCOCO Image Captioning Dataset and WMT14 EN-DE Machine Translation Dataset. 
			Under ``Tokens-Per-Sentence" are the Mean, Standard Deviation, and Median of the Target Sentence Length. }
		\label{tab:dataset}%
		\begin{tabular}{ccccc}
			\toprule
			\multirow{2}[1]{*}{Dataset}  &\multirow{2}[1]{*}{\#Examples} &  \multicolumn{3}{c}{Tokens-Per-Sentence}  \\
			&							   & Mean & StdDev & Median \\
			\midrule
			\midrule
			MSCOCO & 0.6M & 10.5 & 2.4  & 10.0 \\
			WMT14 EN-DE & 4.5M  & 29.4  & 17.0 & 26.0 \\
			\bottomrule
		\end{tabular}%
	\end{table}%

	\section{Experiments on Machine Translation} 
	To validate the generalization ability of our non-autoregressive sequence generation method, we further perform experiments on machine translation task. 
	Compared to image captioning task, machine translation task often has much longer target sentence length (shown in Table \ref{tab:dataset}) and is thus more challenging for non-autoregressive decoding models. 
	We denote our Non-Autoregressive machine Translation models as \textbf{NAT}.

	\subsection{Experimental Settings}

	\subsubsection{WMT14 EN-DE Dataset} 
	We use the widely adopted machine translation dataset, WMT14 En-De\footnote{http://www.statmt.org/wmt14/translation-task}, to evaluate the effectiveness of our proposed method. 
	WMT14 En-De has 4.5 million English-to-German sentence pairs.  
	Following previous practices, we employ \texttt{newstest2013} and \texttt{newstest2014} as the validation and test sets respectively.
	All the data are tokenized and segmented into subwords units using byte-pair encoding (BPE) \cite{sennrich-etal-2016-neural}, leading to a vocabulary of size 40k and is shared for source and target languages. 
	We evaluate the translation quality with BLEU \cite{Papineni2002BLEU}.

	\subsubsection{Implementation Details}
		\label{sec:predictor}
	As is shown in Table \ref{tab:dataset}, WMT EN-DE dataset's target sequence lengths varies greatly and its average length is about 3 times longer than that of MSCOCO image captioning dataset. 
	Therefore, instead of using a fixed number of agents in the decoder, we make the number of agents to adapt to the input sentence. 
	That is, we train a target length predictor to predict the target sequence length and set the number of agents to be the predicted length during inference. 
	During training, we use the ground-truth target sentence length and the length predictor is not used. 
	We formulate the target length prediction as a classification problem, predicting the length offset between the target and source lengths. 
	Specifically, We first perform global pooling on the hidden vectors from the last layer of the encoder, and then feed the resulting vector to a linear layer followed by a softmax function. 
	The length predictor is jointly trained with the whole network during training. 
	
	We use the default network architecture of the Transformer-\textit{Base} \cite{vaswani2017attention} model. 
	We first train an autoregressive Transformer teacher model with standard cross-entropy loss, which achieves a BLEU score of $27.20$. 
	During decoding of the teacher model, beam search with a beam width of 4 is used. 
	Our models are implemented based on the open-sourced \texttt{fairseq} \cite{ott2019fairseq} sequence modeling toolkit. 
	
	For our non-autoregressive models, we first pre-train NAT model with XE loss for 300 epochs,  
	exactly following the Transformer's training settings. 
	The knowledge distillation data produced by the Transformer teacher model is used in this process. 
	After that, we run our CMAL training to optimize the GLEU score \cite{wu2016google} for 30 more epochs with a fixed learning rate of $1\times10^{-5}$. 
	The BLEU score was designed to be a corpus measure and has some undesirable properties when used for calculating per sentence rewards.  
	Therefore, we use a slightly different score, the sentence-level ``GLEU score" \cite{wu2016google}, as our rewards objective. 
	GLEU is designed to be a per-sentence metric while at the same time, correlates quite well with BLEU on a corpus level. 
	To enhance training efficiency, we sample 5 sentences for each input. 
	By default, we use $k=1$ top-ranking words in CMAL. 
	We do not average multiple check-pointed models.

	\begin{table*}[htbp]
		\centering
		\caption{BLEU Score, Latency, and Speedup on Official Test Set of WMT14 En-De Machine Translation Dataset. 
			All of the Compared Non-Autoregressive Models Are Based on Transformer Architecture. 
			The Transformer \cite{vaswani2017attention} Results Are Based on Our Own Reproduction and is Used as the Teacher Model for Knowledge Distillation. 
			``/" Denotes That the Results are Not Reported or Not Directly Comparable. 
			Latency is the Time to Decode a Single Sentence without Minibatching, Averaged Over the Whole Test Split, and is Tested on a Tesla V100 GPU.
			The Speedup Values of the Compared Models are from the Corresponding Papers. 
		}
		\label{tab:nat_main_nmt}%
		\begin{tabular}{lcccc}
			\toprule
			\multirow{2}[1]{*}{Models} &   Decoding & \multicolumn{1}{c}{En$\to$De} & \multirow{2}[1]{*}{Latency} & \multirow{2}[1]{*}{Speedup} \\
			& Iterations & \multicolumn{1}{c}{BLEU}      & 	& \\
			\midrule
			\midrule
			\multicolumn{5}{l}{\textbf{Autoregressive models}} \\
			\midrule
			LSTM-based S2S \cite{wu2016google}  & $N$ & 24.60  & / & / \\
			ConvS2S \cite{pmlr-v70-gehring17a} & $N$ & 26.43  & / & / \\
			Transformer \cite{vaswani2017attention}  & $N$ & 27.20  & 251.4ms & 1.00$\times$ \\
			\midrule 
			\multicolumn{5}{l}{\textbf{Non-autoregressive models}} \\
			\midrule
			NAT with Fertility \cite{gu2017non}		 & 1& 17.69  & / & 15.6$\times$  \\
			LT \cite{pmlr-v80-kaiser18a} & 1 & 19.80  & / & 5.78$\times$ \\
			Iterative Refinement \cite{lee2018deterministic} & 1 & 13.91  & / & 11.39$\times$ \\ 
			Iterative Refinement \cite{lee2018deterministic} & 2 & 16.95 & / & 8.77$\times$ \\ 
			Iterative Refinement \cite{lee2018deterministic} & 10 & 21.61 & / & 2.01$\times$ \\ 
			CTC	\cite{libovicky-helcl-2018-end} & 1 & 17.68  & / & 3.42$\times$ \\
			Auxiliary Regularization \cite{wang2019non}  & 1 &20.65 & / & / \\
			FlowSeq \cite{ma2019flowseq} & 1 & 21.45 & / & / \\
			Bag-of-ngrams Loss \cite{DBLP:conf/aaai/ShaoZFMZ20}  & 1&20.90 & / & 10.77$\times$ \\
			FCL-NAT \cite{guo2020fine} & 1 & 21.70& / & / \\
			\midrule
			\multicolumn{5}{l}{\textbf{Non-autoregressive models (Ours)}} \\
			\midrule
			NAT-XE   &  1 & 18.46 &	\multirow{4}[1]{*}{\bf14.4ms} &  \multirow{4}[1]{*}{\bf17.46$\times$} \\
			NAT-CMAL &  1 & 23.69  &       &  \\
			NAT-CMAL-CA&  1 & 23.87  &       &  \\
			NAT-CMAL-CA (postprocess) &  1 & \bf24.46 &       &  \\
			\bottomrule
		\end{tabular}%
	\end{table*}%

	\subsection{Main Results} 
	We compare our model with the following non-autoregressive methods. 
	NAT with fertility \cite{gu2017non} is the first to propose non-autoregressive neural machine translation, which predicts input token
	fertilities as a latent variable. 
	Latent Transformer (LT) \cite{pmlr-v80-kaiser18a} incorporates an autoregressive module into NAT to predict a sequence of discrete latent variables. 
	Iterative Refinement \cite{lee2018deterministic} trains extra decoders to iteratively refine the translation output with multiple iterations. 
	CTC	\cite{libovicky-helcl-2018-end} uses connectionist temporal classification. 
	Auxiliary Regularization \cite{wang2019non} adds two auxiliary regularization terms on the decoder representations. 
	FlowSeq \cite{ma2019flowseq} introduces a method based on generative flow. 
	Bag-of-ngrams Loss \cite{DBLP:conf/aaai/ShaoZFMZ20} minimizes the bag-of-ngrams difference between the model output and the reference sentence. 
	FCL-NAT \cite{guo2020fine} introduces curriculum learning into fine-tuning of NAT. 
	We also compare with strong autoregressive methods that are based on LSTM \cite{wu2016google}, CNN \cite{pmlr-v70-gehring17a} and self-attention \cite{vaswani2017attention}. 
	The inference latency is the average per-sentence decoding latency over the newstest2014 test set, and is conducted on a single Tesla V100 GPU. 
	The speedup relative to the autoregressive Transformer is also reported.  
	Results are shown in Table \ref{tab:nat_main_nmt}. 
	
	The proposed multi-agent reinforcement learning method with individual counterfactual baseline (NAT-CMAL, Eqn. (\ref{eqn:pg-cf})) outperforms the cross-entropy loss based baseline (NAT-XE, Eqn. (\ref{eqn:xe})) by a significant margin of $5.23$ BLEU points. 
	Considering the influence of compositional agents (CA, Eqn. (\ref{eqn:pg-cf1})) on the counterfactual baseline slightly increases the BLEU score by $0.18$ points. 
	NAT-CMAL-CA achieves state-of-the-art performance with significant improvements over compared non-autoregressive models. 
	Specifically, NAT-CMAL-CA outperforms NAT with fertility by $6.18$ BLEU points and  
	Iterative Refinement with rescoring 10 candidates by $2.26$ BLEU points. 
	Comparing to autoregressive models, NAT-CMAL-CA is only $3.33$ BLEU points behind its Transformer teacher, 
	and is even comparable to the state-of-the-art LSTM-based S2S baseline. 
	By removing the consecutive identical tokens as a simple postprocessing following\cite{lee2018deterministic} (postprocess), the BLEU score of NAT-CMAL-CA is further improved by $0.59$. 
	Our final model shrinks the gap between the autoregressive teacher and the NAT model from $8.74$ ($27.20$ v.s. $18.46$) to only $2.74$ ($27.20$ v.s. $24.46$) BLEU points. 
	The promising results demonstrate that the proposed 
	multi-agent reinforcement learning method can produce higher quality translations 
	by enforcing sentence-level consistency and optimizing the non-differentiable test metric.

	Comparing speedups, our method obtains the greatest speedup of $17.46\times$ over the 
	autoregressive counterpart, with latency significantly reduced from $251.4ms$ to only $14.4ms$.
	Previous works often rely on adding extra components, \eg iterative refinement \cite{lee2018deterministic}, autoregressive submodule \cite{pmlr-v80-kaiser18a}, or CTC decoder \cite{libovicky-helcl-2018-end}, on the Transformer to achieve better generation quality, 
	which, however, inevitably sacrifice the decoding speed. 
	Our method adds no extra components except the lightweight length predictor, and therefore is able to maximize the decoding speedup.

	\subsection{Comparison of Various Reward Baselines $b$} 
\begin{figure}[!t] 
		\centering
		\includegraphics[width=3in]{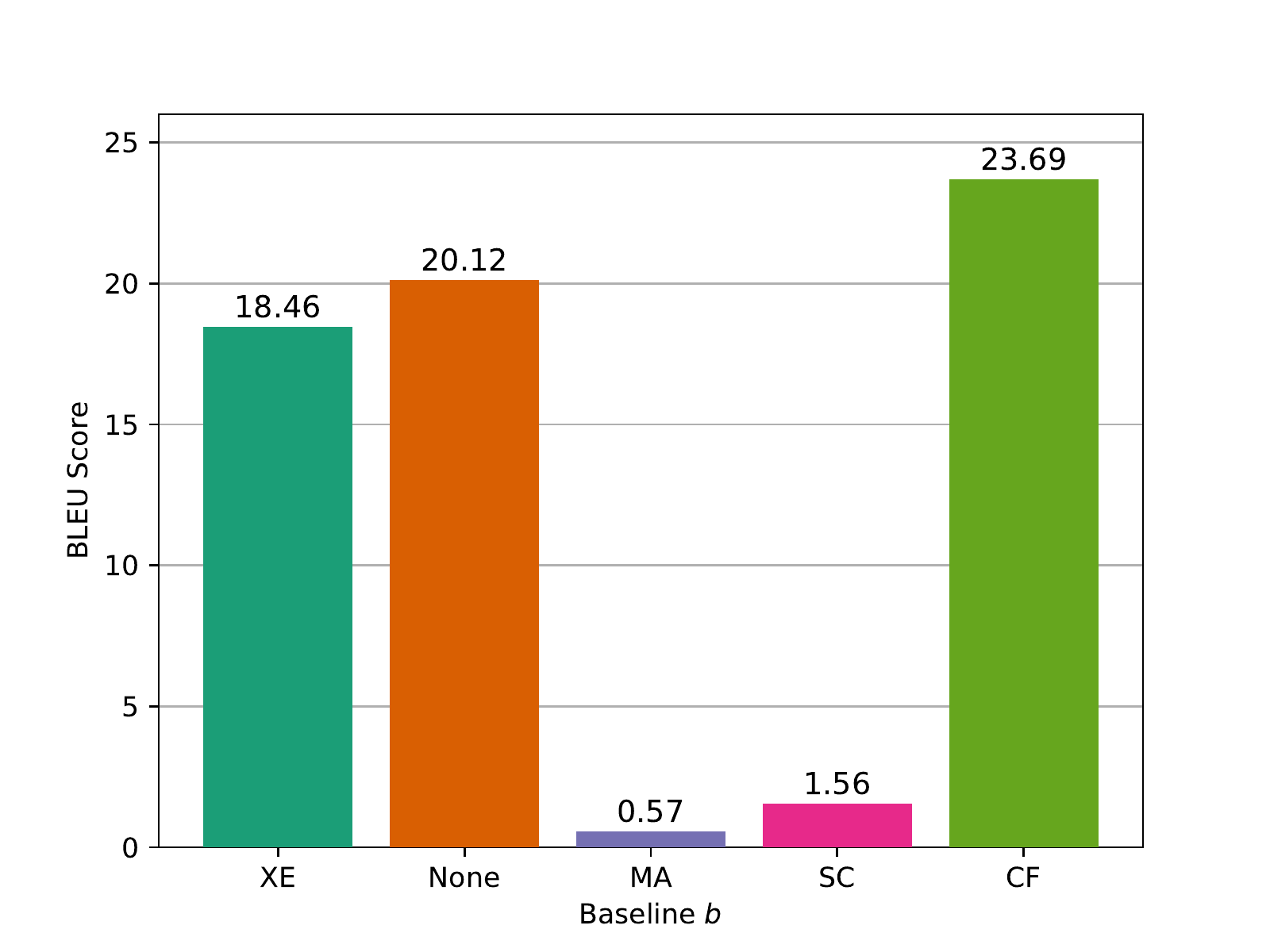}
		\caption{Comparison of using various baselines $b$ (in Eqn. (\ref{eqn:pg}) on WMT14 En-De machine translation dataset.  
			XE: the performance after pre-training with cross-entropy loss. 
			None: not using a baseline. 
			MA: moving average baseline. 
			SC: self-critical baseline. 
			CF: the proposed counterfactual baseline.
		}
		\label{fig:baseline}
	\end{figure}

	Similar to Section \ref{sec:baseline}, we also compare our counterfactual (CF) baseline with two widely-used baselines in policy gradient, 
	\ie Moving Average \cite{weaver2001optimal} (MA) and Self-Critical \cite{Rennie2016Self} (SC), and not using a baseline (\textit{None}), \ie$b=0$. 
	The results are shown in Figure~\ref{fig:baseline}. 
	MA and SC baselines lead to extremely poor performance on WMT14 EN-DE dataset. 
	Not using a baseline (\textit{None}) brings $1.66$ BLEU improvements over the XE model. 
	Our CF baseline significantly outperforms all the other compared baselines by large margins. 
	Particularly, CF outperforms \textit{None} by $3.57$ BLEU points.  
	The results demonstrate that the proposed counterfactual baseline can efficiently assign credit to each agent and thus stabilize the training process.

	\subsection{Effect of Top-$k$ Size}
	As shown in Table~\ref{tab:topk_nmt}, the model is insensitive to top-$k$ size.  
	Simply setting $k$ to 1 could achieve fairly good performance.

	\begin{figure}[!tp] 
		\centering
		\includegraphics[width=3in]{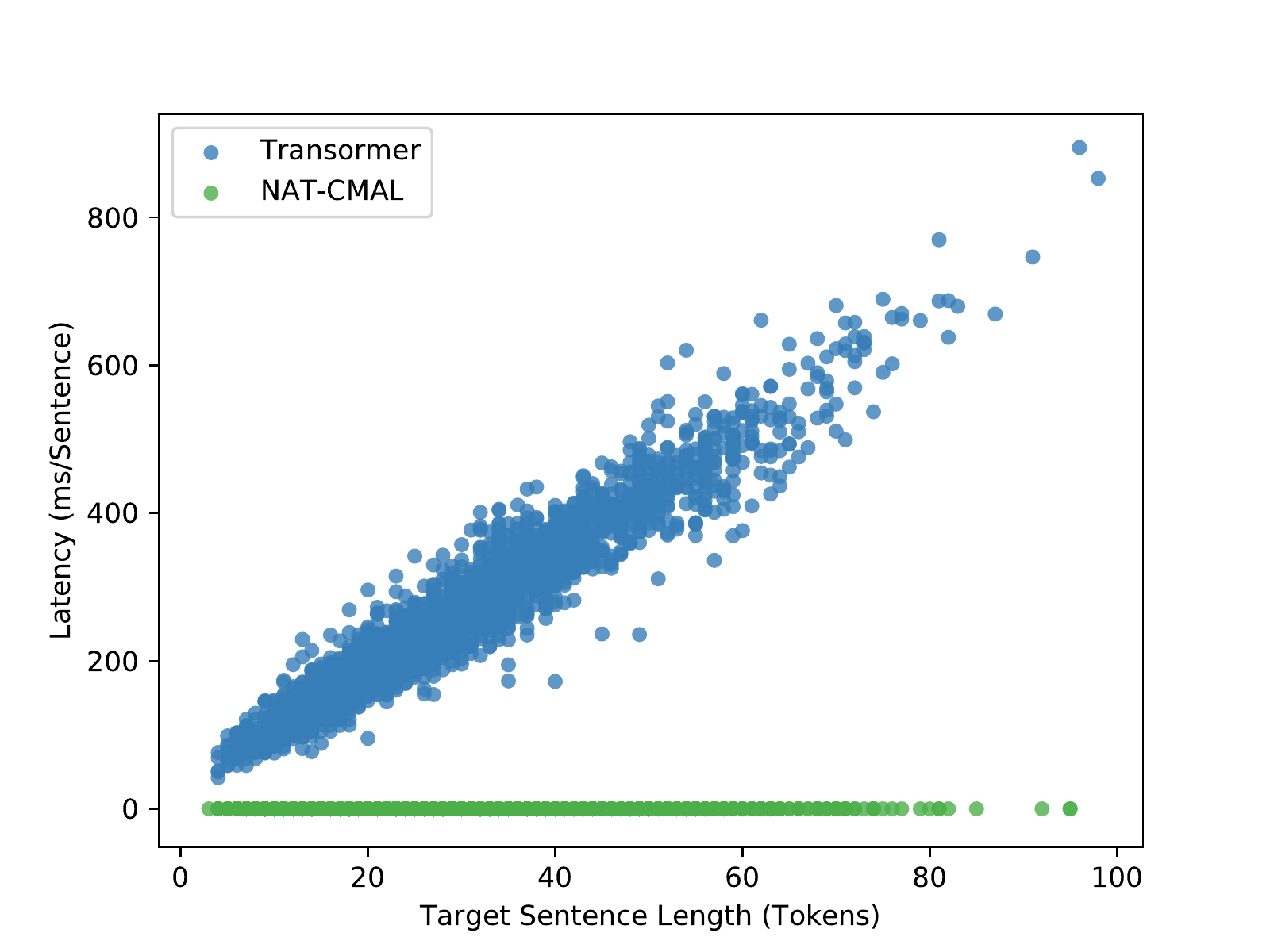}
		\caption{
			The translation latencies for each sentence in the official test set of WMT14 En-De machine translation dataset. 
			Latency is the time used to decode a single sentence without minibatching, 
			averaged over three separate runs, and is tested on a GeForce GTX 1080 Ti GPU.
		}
		\label{fig:length_latency}
	\end{figure}

	\subsection{The Effects of Sentence Lengths} 
\begin{figure}[!tp] 
		\centering
		\includegraphics[width=3in]{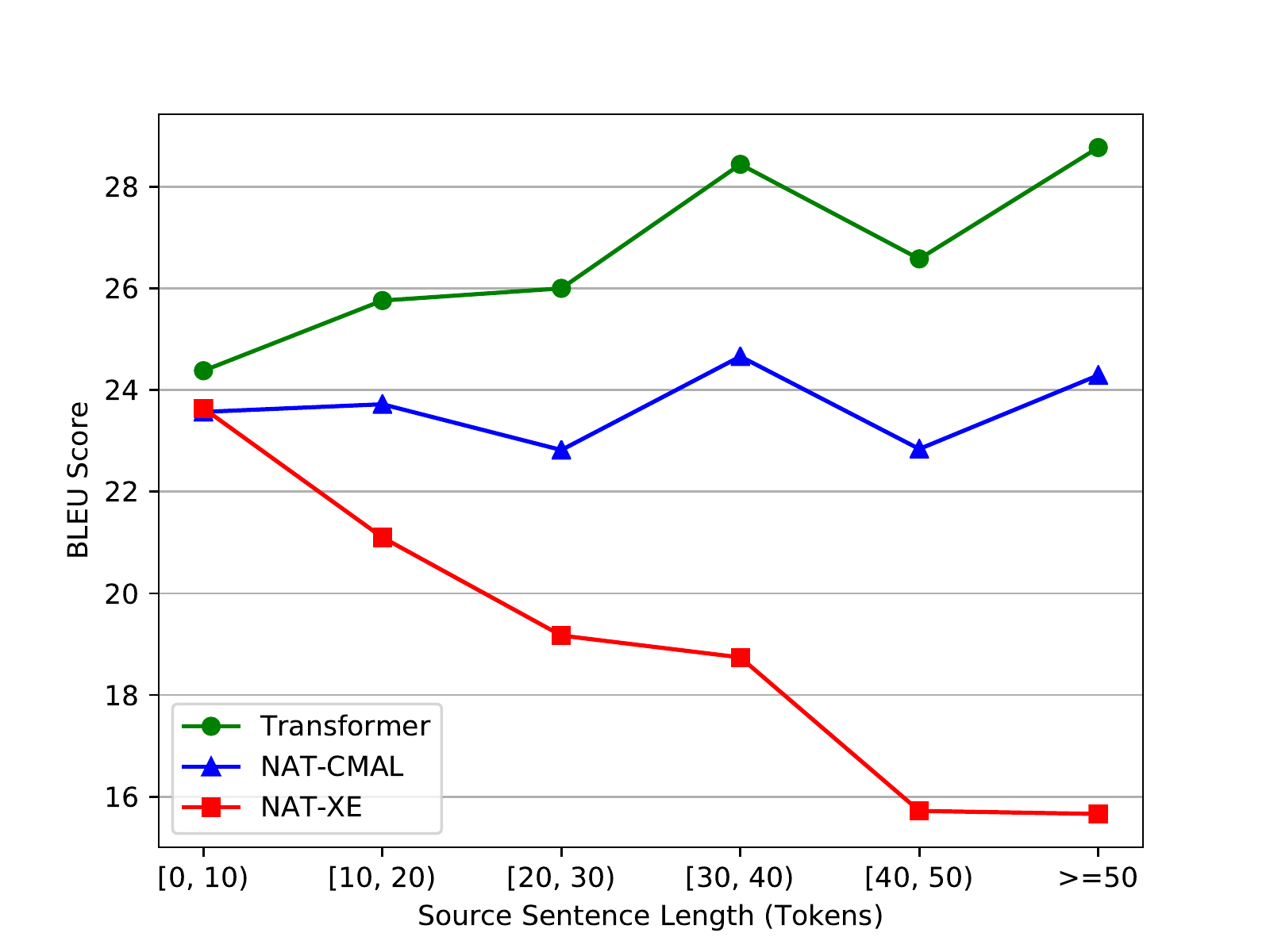}
		\caption{
			The BLEU scores comparison between Transformer, NAT-XE,
			and our NAT-CMAL over sentences in different length buckets on
			official test set of WMT14 En-De machine translation dataset. 
		}
		\label{fig:sentence_length}
	\end{figure}

	We compare the translation quality between Transformer \cite{vaswani2017attention}, NAT-XE and the proposed NAT-CMAL with regard to different sentence lengths. 
	We divide the sentence pairs in WMT14 En-De test set into different length buckets according to the length of source sentence (number of subword tokens).
	The results are shown in Figure \ref{fig:sentence_length}. 
	It can be seen that as sentence 
	length increases, the accuracy of NAT-XE drops quickly and the gap between Transformer and NAT-XE also enlarges. 
	Our NAT-CMAL has stable performance over various sentence lengths, and in particular, performs much better than NAT-XE over long sentences. 
	The performance degradation of NAT-XE is because the drawback of word-level cross-entropy loss, \ie cannot model sentence-level consistency, magnifies as sentence length grows. 
	Our CMAL method optimizes the sentence-level objective with cooperative agents and is therefore better at maintaining the global consistency of long sentences.

	In Figure \ref{fig:length_latency}, we compare the per-sentence decoding latency between the autoregressive Transformer and the proposed NAT-CMAL. 
	We conduct the analysis on WMT14 En-De test set without minibatching and report the average latency of three separate runs. 
	We can see that the latency of Transformer is linear in the decoding length, 
	while that of our NAT-CMAL is nearly constant for various lengths. 
	When the target sentence length is close to $100$, Transformer requires a decoding time close to 1 second even on a high-performance Tesla V100 GPU, 
	while our NAT-CMAL only uses $16.4ms$ to decode the sentence.  
	
		\begin{table}[tp]
		\centering
		\caption{Effect of Top-$k$ Size (in Eqn. (\ref{eqn:topk}) on Official Test Set of WMT14 En-De Translation Dataset. }
		\label{tab:topk_nmt}%
		\begin{tabular}{cccc}
			\toprule
			\multicolumn{1}{c}{top-$k$} &1 & 2 & 5 \\
			\midrule
			\midrule
			BLEU & \bf23.69  & 23.67  & 23.66  \\
			\bottomrule
		\end{tabular}%
	\end{table}%

	\subsection{Qualitative Analysis}
	In Table \ref{tab:nmt_case}, we provide two examples of translations from Transformer, NAT-XE, NAT-CMAL, as well as the ground-truth translation. 
	Consecutive identical words, highlighted in gray, are most prevalent in the translations of NAT-XE model, especially for the relatively complex second example sentence. 
	Translations from NAT-CMAL are often more consistent than NAT-XE, and are of similar quality to the outputs produced by the autoregressive Transformer teacher. 
	The examples suggest that CMAL training is better at maintaining the global consistency of sentences than XE loss.

	\section{Conclusion} 
	We have proposed a non-autoregressive sequence generation model and a novel Counterfactuals-critical Multi-Agent Learning (CMAL) algorithm. 
	The decoding inconsistency problem in non-autoregressive models 
	is well addressed by the combined effect of the cooperative agents, sentence-level team reward, and individual/compositional counterfactual baselines. 
	The generation quality is further boosted by augmenting training data with unlabeled inputs.  
	Extensive experiments on MSCOCO image captioning benchmark and WMT14 EN-DE machine translation dataset have shown that our non-autoregressive model significantly outperforms previous cross-entropy trained non-autoregressive models while at the same time enjoys the greatest decoding speedup. 
	In particular, our non-autoregressive image captioning model even achieves a performance very comparable to state-of-the-art autoregressive models. 
	
	There are two promising future directions. 
	First, we can apply our method to more sequence generation tasks, \eg speech recognition and dialog response generation. 
	Second, it will be interesting to extend our method for generating very long sequences (\eg hundreds of words) in non-autoregressive manner. 
	For example, we can design a hierarchical reinforcement learning system, in which a level of manager agents assign goals for their worker agents 
	and the worker agents simultaneously perform actions to generate the whole sequence.

	\ifCLASSOPTIONcompsoc
	\section*{Acknowledgments}
	\else
	\section*{Acknowledgment}
	\fi

	This work was supported by the National Key Research and Development Program of China (No. 2020AAA0106400), National Natural Science Foundation of China (61922086, 61872366), and Beijing Natural Science Foundation (4192059).

	\ifCLASSOPTIONcaptionsoff
	\newpage
	\fi

	\input{naic.bbl}

	\begin{IEEEbiography}[{\includegraphics[width=1in,height=1.25in,clip,keepaspectratio]{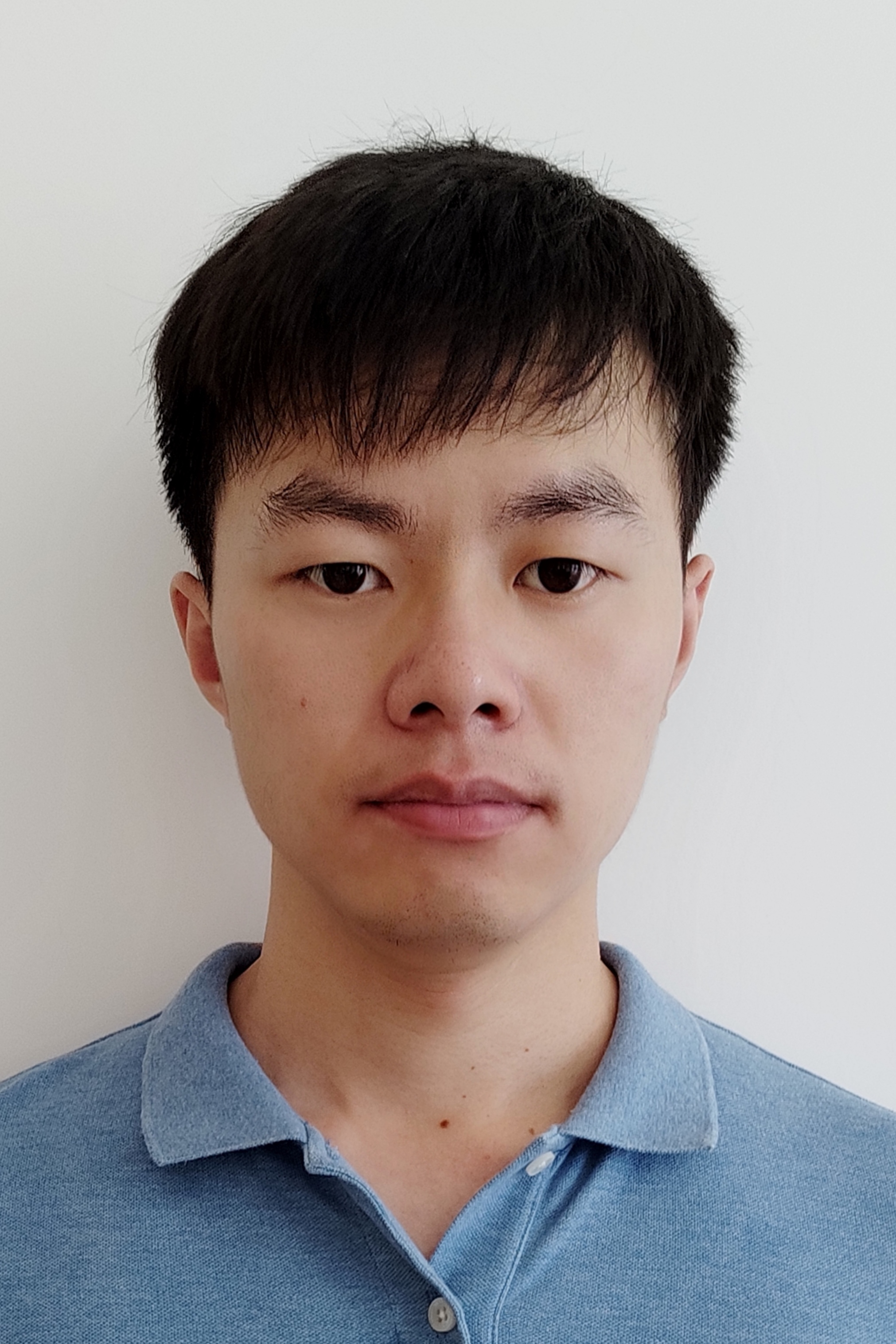}}]{Longteng Guo}
		received the B.E. degree from Xi'an Jiaotong University, Shaanxi, China, in 2016. He is currently pursuing the Ph.D. degree at National Laboratory of Pattern Recognition, Institute of Automation, Chinese Academy of Sciences, Beijing, China. His current research interests include deep learning, image content analysis and multimodal deep learning.
	\end{IEEEbiography}
	
	\begin{IEEEbiography}[{\includegraphics[width=1in,height=1.25in,clip,keepaspectratio]{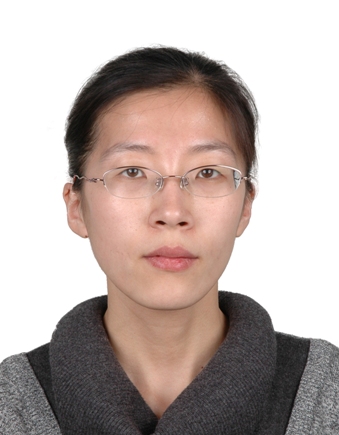}}]{Jing Liu}
		received the Ph.D. degree from the Institute of Automation, Chinese Academy of Sciences, Beijing, in 2008. She is a Professor with the National Laboratory of Pattern Recognition, Institute of Automation, Chinese Academy of Sciences. Her current research interests include deep learning, image content analysis and classification, multimedia understanding and retrieval. 
	\end{IEEEbiography}
	
	\begin{IEEEbiography}[{\includegraphics[width=1in,height=1.25in,clip,keepaspectratio]{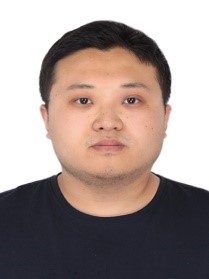}}]{Xinxin Zhu}
		Xinxin Zhu received the B.E. degree from Hebei Normal University, Hebei, China, in 2013, and the Ph.D. degree from Beijing University of Posts and Telecommunications, Beijing, China, in 2019. He is an assistant professor with the National Laboratory of Pattern Recognition, Institute of Automation, Chinese Academy of Sciences. His current research interests include deep learning, image content analysis and multimodal deep learning.
	\end{IEEEbiography}

	\begin{IEEEbiography}[{\includegraphics[width=1in,height=1.25in,clip,keepaspectratio]{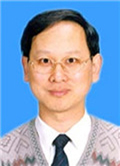}}]{Hanqing Lu}
		received the Ph.D. from Department of Electronic and Information Science in Huazhong University of Science and Technology. He is a Professor with the National Laboratory of Pattern Recognition, Institute of Automation, Chinese Academy of Sciences. His current research interests include image similarity measure, video analysis, multimedia technology and system.
	\end{IEEEbiography}

\end{document}